\documentclass{article}

   \usepackage[nonatbib, preprint]{neurips_2023}
\usepackage[numbers, square, sort&compress]{natbib}

\usepackage[utf8]{inputenc} % allow utf-8 input
\usepackage[T1]{fontenc}    % use 8-bit T1 fonts
\usepackage{hyperref}       % hyperlinks
\usepackage{adjustbox}
\usepackage[most]{tcolorbox}

\usepackage{url}            % simple URL typesetting
\usepackage{booktabs}       % professional-quality tables
\usepackage{amsfonts}       % blackboard math symbols
\usepackage{nicefrac}       % compact symbols for 1/2, etc.
\usepackage{microtype}      % microtypography
\usepackage{graphicx}
\usepackage{subcaption}
\usepackage{amsthm}
\theoremstyle{definition}

\newcommand*{\myfont}{\fontfamily{cmr}\selectfont}
\title{Interpretable Medical Diagnostics with Structured Data Extraction by Large Language Models}

\author{%
  Aleksa Bisercic \\
  University Singidunum\\
  \texttt{aleksa.bisercic.22@singimail.rs} \\
  \And
  Mladen Nikolic \\
  University of Belgrade\\
  \texttt{mladen.nikolic@matf.bg.ac.rs} \\
  \And
  Mihaela van der Schaar \\
  University of Cambridge,\\
  The Alan Turing Institute \\
  \texttt{mv472@cam.ac.uk} \\
  \And
  Boris Delibasic \\
  University of Belgrade\\
  \texttt{boris.delibasic@fon.bg.ac.rs} \\
  \And 
  Pietro Lio \\
  University of Cambridge\\
  \texttt{pl219@cam.ac.uk} \\
    \And
  Andrija Petrovic \\
  University of Belgrade\\
  \texttt{andrija.petrovic@fon.bg.ac.rs} \\
}

\begin{document}

\maketitle

\begin{abstract}
Tabular data is often hidden in text, particularly in medical diagnostic reports. Traditional machine learning (ML) models designed to work with tabular data, cannot effectively process information in such form. On the other hand, large language models (LLMs) which excel at textual tasks, are probably not the best tool for modeling tabular data. Therefore, we propose a novel, simple, and effective methodology for extracting structured tabular data from textual medical reports, called TEMED-LLM. Drawing upon the reasoning capabilities of LLMs, TEMED-LLM goes beyond traditional extraction techniques, accurately inferring tabular features, even when their names are not explicitly mentioned in the text. This is achieved by combining domain-specific reasoning guidelines with a proposed data validation and reasoning correction feedback loop. By applying interpretable ML models such as decision trees and logistic regression over the extracted and validated data, we obtain end-to-end interpretable predictions. We demonstrate that our approach significantly outperforms state-of-the-art text classification models in medical diagnostics. Given its predictive performance, simplicity, and interpretability, TEMED-LLM underscores the potential of leveraging LLMs to improve the performance and trustworthiness of ML models in medical applications.
\end{abstract}

\section{Intorduction}

Tabular data, consisting of numerical values arranged in rows and columns is prevalent in numerous real-world scenarios~\citep {benjelloun2020google}. It appears in a wide array of practical applications spanning various industries, such as healthcare~\citep{qayyum2020secure}, finance~\citep{ghoddusi2019machine}, and many others~\citep{angra2017machine,dogan2021machine,tahmasebi2020machine}. This led to the development of numerous methods aimed at extracting valuable insights from this kind of data~\citep{shwartz2022tabular, borisov2022deep}, ultimately driving advancements in diverse fields and facilitating informed decision-making. 

In numerous health and medical problems, the underlying data is often of a tabular nature, albeit obscured by a textual form such as doctor reports, patient records, or clinical notes. For processing this kind of data text-based machine learning (ML) methods are employed~\citep{mullenbach2018explainable, locke2021natural, liu2020natural}. There are several key reasons to prefer tabular ML methods over text-based ML methods in the case when the true underlying problem is of tabular nature. \textbf{Efficiency.} Thanks to its simple but precise structure, tabular data often allows for computationally and statistically more efficient ML algorithms than the textual data~\citep{shwartz2022tabular,tripathy2021comprehensive}. Moreover, textual data requires additional preprocessing steps, such as tokenization, stemming, and lemmatization, which complicate the processing pipeline. \textbf{Interpretability.} Tabular data usually consists of well-defined, human-interpretable features. This allows some ML algorithms to capture the relevant patterns through interpretable tabular models~\citep{du2019techniques}. In contrast, textual models are often large and complex and do not admit direct interpretation~\citep{gholizadeh2021model}. \textbf{Applicability.} ML algorithms for tabular data are applicable across various domains due to the structured nature of the data. Conversely, textual data models may require significant adjustments and fine-tuning when applied to different domains, as language patterns, context, and vocabulary can vary greatly. 

Inspired by the latest accomplishments of large language models (LLMs)~\citep{sallam2023utility} in generating synthetic tabular data~\citep{borisov2022language}, few-shot learning for tabular datasets~\citep{hegselmann2023tabllm}, and reasoning~\citep{yao2022react,huang2022towards} we propose a methodology of Tabular Extraction for Enhanced Medical Diagnostics using Large Language Models (TEMED-LLM). TEDEM-LLM utilizes LLMs in the extraction of tabular data from medical texts and reports based on predefined features. To do so, it relies on reasoning guidelines and a new data validation and reasoning correction feedback loop. The extracted data is then processed using interpretable ML models for tabular data such as decision trees and logistic regression, resulting in end-to-end interpretable predictions. Our approach outperforms state-of-the-art text classification models in several medical diagnostics tasks while offering the simplicity of modeling and interpretability of obtained models and decisions. Our evaluation is conducted both on publicly available datasets and on electronic health records obtained from clinicians. We hope this research boosts decision-makers' confidence in employing ML models in medical diagnostics. 

\iffalse
\begin{figure}
	\center
	\includegraphics[angle=0, width=0.8\textwidth]{images/Introduction.png}
	\captionsetup{justification=centering}
	\caption{Rough workflow sketch illustrating the conversion of medical and clinical reports into structured data tables and the subsequent application of tabular data machine learning model.}
	\label{fig:Fig1}
\end{figure}
\fi

\textbf{Contributions.} 
The contributions of this paper are the following: i) a novel, simple, and effective approach for extracting structured tabular data from textual medical reports; ii) a data validation and reasoning correction feedback loop that improves extraction quality; iii) a significant boost in predictive power over state-of-the-art text classification models in medical diagnostics tasks; iv) interpretable modeling pipeline.

%The remainder of the paper is structured as follows. In section~\ref{Sec:Related Work} the related work is reviewed. TEMED-LLM model architecture is presented in section~\ref{Sec:Methodology}. Experimental setup and results on real-world datasets are presented in section~\ref{Sec:exp-evaluation}. The conclusions are drawn in section~\ref{Sec:Conclusion}.

\section{Challenges}
\label{Sec:Challenges}
Clinicians face numerous challenges in the application of ML methods to the problem of diagnostic prediction. Some of these challenges are related to the complexities of the data extraction for tabular ML methods such as: \textbf{medical terminology and jargon} - medical reports are filled with specialized terminology and abbreviations that can be challenging to interpret, especially when they are used inconsistently or in a context-specific manner; \textbf{complex sentence structures} - the complex sentence structures often found in medical reports can make it difficult to extract information accurately; \textbf{ambiguity and uncertainty} - medical reports can contain ambiguous or uncertain information, such as vague descriptions of symptoms or unclear diagnostic results; \textbf{contextual understanding} - medical data often requires a deep understanding of the patient’s history and context to be interpreted correctly; \textbf{human effort} - extracting features from medical reports manually is time-consuming and can be done only by experienced clinicians. Tabular ML methods cannot easily avoid these issues, but modern LLMs seem well-positioned to offer a satisfactory solution.

One might apply textual ML methods directly, but face different challenges such as: \textbf{integration of multimodal data} - medical reports often come with complementary data of tabular nature, such as various lab or screening results, which do not constitute text and are not suitable for text-based NLP algorithms; \textbf{transparency and interpretability} - it is crucial for clinicians to understand why an NLP algorithm made a particular prediction or diagnosis, but advanced NLP models are mostly black boxes. Interpretable tabular ML methods offer obvious advantages in the context of these issues.

TEMED-LLM provides the best of both worlds. It addresses the challenges of data extraction by leveraging LLM technology and avoids the challenges related to textual modeling by relying on tabular data.

\section{Related work}
\label{Sec:Related Work}

\textbf{ML for tabular medical data analysis.}
ML has revolutionized the field of healthcare by offering novel ways to analyze tabular medical data~\citep{shailaja2018machine, bhardwaj2017study}, which typically comprises structured information from electronic health records~\citep{shickel2017deep}, clinical trials~\citep{zame2020machine}, and patient registries~\citep{lin2020evaluation}. 
%Healthcare related data is often of tabular nature, due to doctors' domain knowledge that informs them of the key features influencing potential patient outcomes and diagnoses.
By employing techniques such as classification, regression, and clustering, ML models can identify underlying patterns and associations~\citep{jones2022artificial}, predict patient outcomes~\citep{alaa2019attentive}, and personalize treatment plans~\citep{alaa2017bayesian}. With the rapid advancement of ML algorithms and the increasing availability of large-scale medical datasets~\citep{dash2019big}, the potential for transforming healthcare through tabular medical data analysis continues to expand. An especially important trend is that of developing interpretable ML models in healthcare~\citep{ahmad2018interpretable}, as it is crucial for doctors to understand the decision-making process behind these models to ensure trust and effective clinical application~\citep{rudin2019stop}. 

\textbf{Medical diagnostics and prediction by NLP.}
There is a growing interest in leveraging unstructured medical data, such as doctors' reports, clinical notes, and medical texts, to predict patient outcomes and diagnoses using ML techniques~\citep{locke2021natural}. Studies such as~\citep{locke2021natural} highlight the utility of textual inputs for predicting patient states. A wide variety of natural language processing (NLP) models has been developed for this purpose, including attentional convolutional networks that predict medical codes from clinical texts, as demonstrated in~\citep{mullenbach2018explainable}.
Moreover, numerous recent studies have explored the use of state-of-the-art transformer-based frameworks for predicting disease progression risk from unstructured clinical notes. These frameworks include BERT~\citep{mao2022ad}, hierarchical multimodal BERT~\citep{agarwal2021preparing}, BioBERT~\citep{lee2020biobert}, and LLMs~\citep{chen2023boosting}. However, one of the major challenges associated with transformer-based models in medical prediction tasks is their limited explainability and interpretability, as noted by~\citep{amjad2023attention}. As the demand for interpretable ML models in healthcare grows, this paper addresses the challenge by developing a simple yet efficient model that can enhance doctors' and clinicians' trust in using ML models for patient outcome prediction and diagnosis.

\textbf{LLMs for tabular data.}
The recent development of LLMs has generated considerable excitement due to their ability to handle a wide variety of tasks, demonstrating impressive performance in natural language understanding and generation~\citep{tamkin2021understanding, boiko2023emergent}. Studies have shown that incorporating reasoning in LLMs can significantly improve their performance across various tasks, including question-answering and commonsense reasoning~\citep{yao2022react, huang2022towards}. Despite these advances, there is still limited research on the application of LLMs to tabular data. A novel entity matching technique based on pre-trained language models is presented in~\citep{li2020deep}, showcasing the potential of LLMs in handling structured data. Recently it has been demonstrated that LLMs can be successfully applied to generate synthetic tabular data~\citep{borisov2022language}. The proposed method -- GReaT (Generation of Realistic Tabular data with LLM), is a pioneering model that leverages an auto-regressive generative LLM to sample synthetic and highly realistic tabular data. Importantly, GReaT can model tabular data distributions and condition on any subset of features, thus providing flexibility in data generation tasks.
In the context of few-shot learning with LLMs, researchers have developed an innovative approach -- TabLLM~\citep{hegselmann2023tabllm}, which involves serializing the tabular data into a natural-language string, along with a brief description of the classification problem. This simple yet effective model outperforms several tabular data few-shot learners on benchmark datasets. 
Building on these successful results and inspired by the capabilities of LLMs, we have incorporated LLMs as one of the key building blocks in TEMED-LLM. In comparison to the GReaT, we do not work with synthetically generated data but employ LLMs to extract tabular data from real-world texts. In contrast to TabLLM, we use LLMs to extract tabular data from text, instead of encoding tabular data in the text and classifying it by a LLM.

\section{TEMED-LLM: Medical Text to Tabular Data Conversion with Improved Interpretability and Diagnostic Performance}
\label{Sec:Methodology}

\begin{figure}
	\center
	\includegraphics[angle=0, width=1\textwidth]{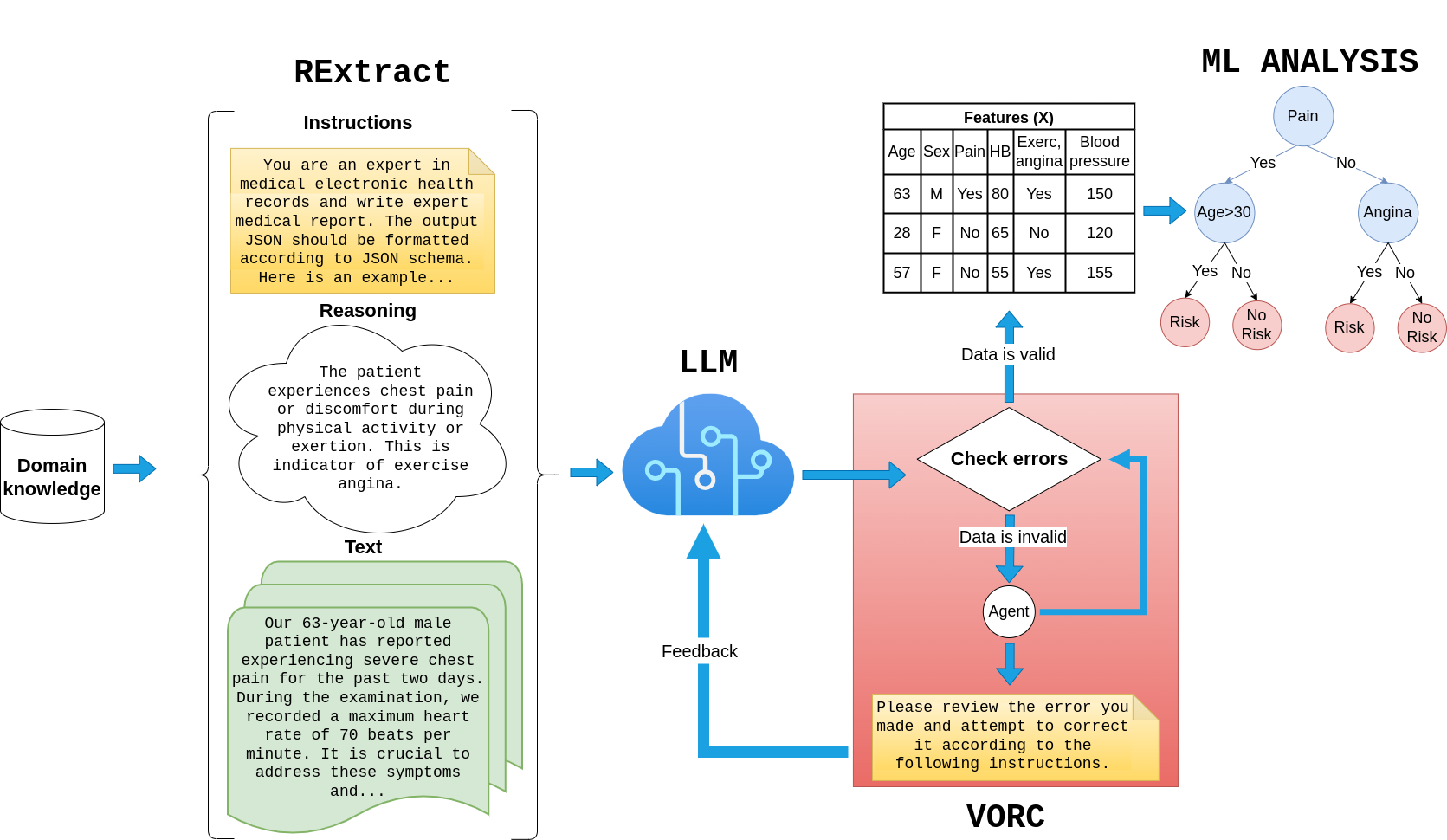}
	\captionsetup{justification=centering}
	\caption{A schematic representation of the TEMED-LLM architecture, featuring the {\myfont RExtract} module for generating LLM prompts using domain knowledge, reasoning, and few-shot examples; the {\myfont VORC} module for error recognition and correction through agent assistance or correction prompts; and the ML analysis component for training ML models on extracted tabular data.}
	\label{fig:Fig2}
\end{figure}

We present TEMED-LLM, a methodology that leverages an LLM to transform the medical text into tabular data and enhance interpretability and diagnostic performance. In Fig.~\ref{fig:Fig2} a sketch of the TEMED-LLM methodology is given, highlighting its three main components. The first component is the {\myfont RExtract} (Reasoning and Extraction) module, which generates a prompt for the LLM to extract tabular data, incorporating instructions on how to extract attributes, which attributes to extract, few-shot examples, and reasoning. All these instructions are defined using domain knowledge. The second component is the {\myfont VORC} (Validation Of Result and Reasoning Correction) module, which recognizes errors and corrects them either by employing a rule-based agent or by sending a correction prompt to the LLM. The last component depicted is the ML analysis module, which is responsible for training (preferably) an interpretable ML model from the extracted tabular data. We begin by defining the problem at hand and then proceed to discuss each component in detail.

\subsection{Problem formulation}

The TEMED-LLM methodology addresses two separate challenges: converting a text to a table and predicting labels based on a table. The first one is a problem of finding the mapping $f:{\cal T}\rightarrow{\cal X}$ which maps texts from a set of medical texts ${\cal T}$ to feature vectors from an $m$-dimensional vectorial feature space ${\cal X}$. A set of texts $\{\mathbf{t}_1, \mathbf{t}_2, \ldots, \mathbf{t}_N\}$ is then converted to table $X$, such that for each text $\mathbf{t}_i$ it holds $f(\mathbf{t}_i)=\mathbf{x}_i$. This role is performed by {\myfont RExtract} and {\myfont VORC} modules. The second problem is a problem of finding a function $g:{\cal X}\rightarrow{\cal Y}$ which maps from the feature space to the label space ${\cal Y}$, so that for each vector $\mathbf{x}_i$ it holds $f(\mathbf{x}_i)=y_i$. This is a classical supervised ML task and is performed by the ML analysis module. In our experiments, labels $y_i$ which correspond to texts $t_i$ and vectors $x_i$ are already provided in the datasets. However, if the labels are not provided explicitly, but are contained in medical reports, they can be extracted from them just like the other features.

\subsection{{\myfont RExtract} module}

The prompt generation process plays a vital role in the TEMED-LLM methodology, as crafting precise and effective prompts helps LLM extract accurate and relevant information from input medical texts. Inspired by the work in~\citep{yao2022react}, we introduce a novel prompt generation module called {\myfont RExtract} (Reason and Extract), which focuses on reasoning and extracting data from unstructured medical texts. The input to this module consists of three main components which we discuss in turn.

\textbf{Instructions}: This component instructs the LLM on which task it needs to perform. It is subdivided into three distinct parts. The first part provides general instructions about the feature extraction task and the desired structure of the JSON output file. The second part specifies which features should be extracted, the feature descriptions, and the type of each feature value. The third part provides an example of a medical report along with the ground truth values for features extracted from that report (one-shot example), drawing inspiration from the work presented in~\citep{kojima2022large}.\\
\textbf{Reasoning}: The reasoning component outlines explicit guidance on how the model should reason about the provided example in the instructions. This guidance is provided for each feature that needs to be extracted, ensuring that the LLM possesses a comprehensive understanding of the extraction process and the underlying logic. The reasoning, along with the instructions, should be derived from domain knowledge. Moreover, it is possible to provide different reasoning for different types of medical reports, allowing for greater adaptability to various report formats.\\
\textbf{Text}: This component contains the medical text from which the LLM should extract tabular data.

\begin{figure}
	\center
	\includegraphics[angle=0, width=1\textwidth]{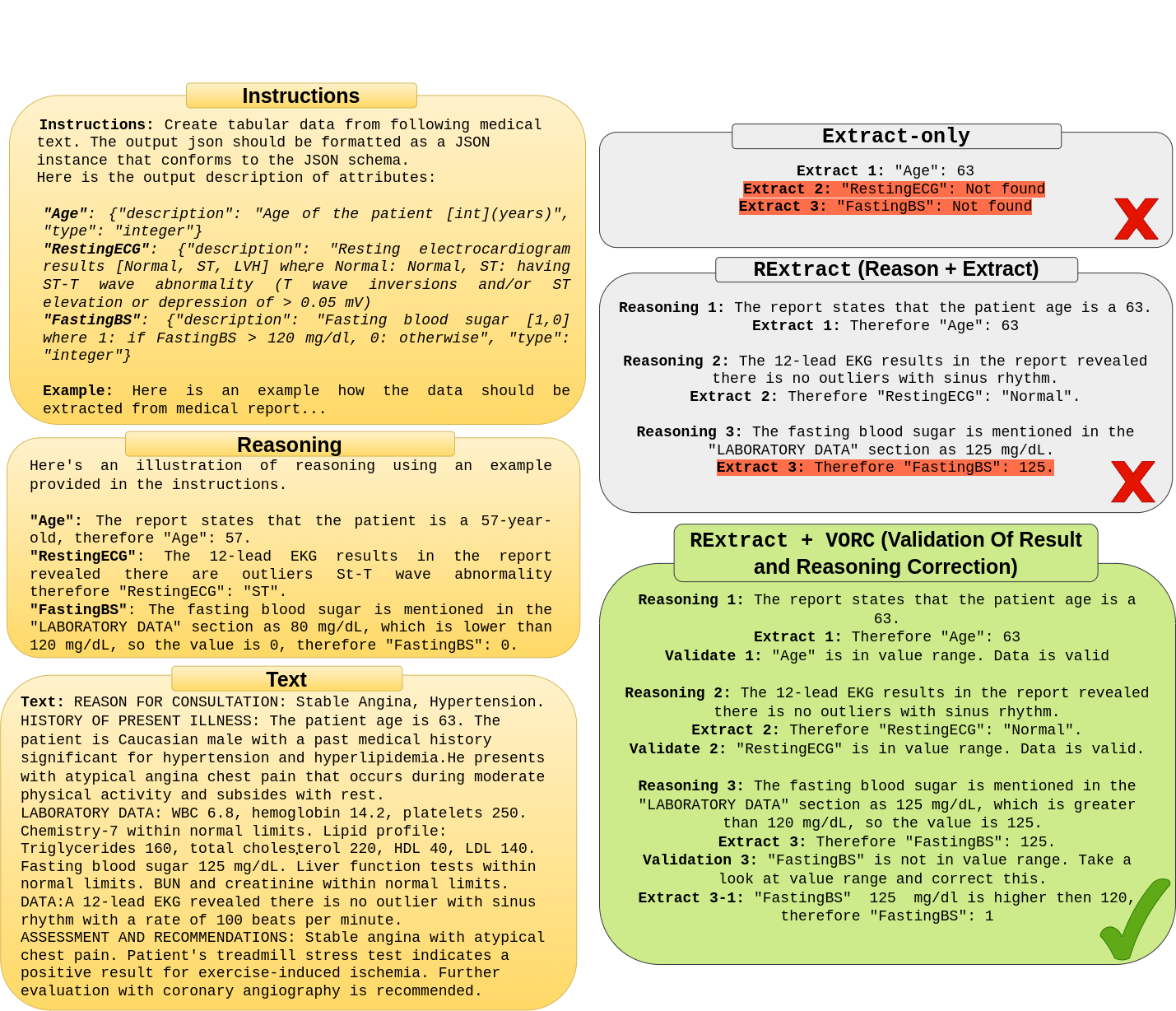}
	\captionsetup{justification=centering}
	\caption{Illustration of tabular data extraction from medical reports. In the left part, the diagram shows the extraction instructions, the reasoning step, and the text from which features should be extracted. The upper right corner emphasizes the shortcomings of {\myfont Extract-only} (Instructions + Text). The middle right illustrates {\myfont RExtract}'s (Instructions + Reasoning + Text) efficiency in recognizing features amidst occasional formatting errors, and the lower right demonstrates the successful collaboration of {\myfont RExtract} and {\myfont VORC}, showcasing the accurate extraction of feature values.}
	\label{fig:Fig3}
\end{figure}

Our methodology is illustrated in Fig.~\ref{fig:Fig3}. The left part presents the instructions, reasoning, and text. The upper right depicts the functioning of an {\myfont Extract-only}(Instructions + Text) module -- a model without reasoning capabilities that relies solely on searching specific feature names from text without using reasoning. In this case, the model struggles with even simple examples and fails to produce accurate results because the feature names cannot be explicitly found in the medical report. The middle right shows the {\myfont RExtract} (Instructions + Reasoning + Text) module in action. This module, equipped with the capacity to identify all pertinent features through reasoning, might sometimes struggle to format feature values accurately. This suggests that the LLM may occasionally present some features in an unsuitable output format. Despite these occasional errors, {\myfont RExtract} is a significant improvement over the extraction-only model, demonstrating the value of incorporating reasoning into the extraction process. The new validation and reasoning correction module -- {\myfont VORC}, is presented in the lower right corner.

\subsection{{\myfont VORC} module}

The data validation procedure in TEMED-LLM is essential for ensuring the accuracy and reliability of information extracted from medical texts. We introduce the {\myfont VORC} module, which detects and corrects errors that may occur during the extraction process, such as JSON validation errors or typing validation errors. The {\myfont VORC} module is a loop in which a user-coded rule-based agent attempts to detect discrepancies of extracted data from feature specifications and to correct minor errors. If unsuccessful, the agent generates a prompt that informs the LLM about the mistake and instructs it to correct the error, thus transitioning to the next loop iteration. The {\myfont VORC} module consists of the following components.

\textbf{JSON error correction}: When the original extraction prompt produces an answer but encounters a JSON validation error, the agent attempts to fix simple errors (e.g., double quotes or single quotes) in the JSON using predefined rules. If the error persists, the LLM is provided with the original prompt, response, and error, and instructed to extract the JSON data once more.\\
\textbf{Key type error correction}: When a key type error is detected, the LLM is informed of the error and instructed to make the necessary corrections.

\subsection{ML analysis module}

The extracted data is modeled by tabular ML techniques. We emphasize the use of interpretable models like decision trees and logistic regression in order to provide informed decision-making. In cases where these simple models cannot achieve satisfactory prediction performance, we propose employing more complex and powerful techniques, such as XGBoost ({\myfont xgboost}). While {\myfont xgboost} model is not equally explainable as logistic regression and decision trees, there is a wide array of explanation techniques that can aid in interpreting its decisions~\cite{molnar2022}.

\section{Experimental evaluation}
\label{Sec:exp-evaluation}

\textbf{Datasets.} We evaluate TEMED-LLM using a variety of real-world medical datasets, created in collaboration with clinicians. All datasets were split into $70\%$ for training, $10\%$ for validation, and $20\%$ for testing. Here we provide short descriptions of the datasets used, whereas more detailed descriptions are provided in appendix~\ref{app:dataset}. The datasets are divided into two groups. The first group consists of datasets of textual medical reports from which the clinicians extracted the tabular data. Datasets of this kind are the following.\\
\textit{Patient treatment classification}: This dataset consists of 718 Electronic Health Records (EHRs) from a private hospital that are described by 10 predefined features. The aim is to determine the course of action for a patient's treatment, whether in-care or out-care, based on lab test results and doctors' observations. \\
\textit{Heart disease diagnosis}: Te dataset includes 917 doctors' EHRs from a private hospital which are described by 11 predefined features. The notes detail patient history, clinical findings, diagnostic test results, pre-and postoperative care, patient progress, and medication. The objective is to identify whether patients have been diagnosed with heart disease.

The second group consists of the publicly available datasets initially in tabular form based on which the clinicians created textual medical reports. These are the following.\\
\textit{Hepatitis prediction}: The dataset (\href{https://www.kaggle.com/datasets/fedesoriano/hepatitis-c-dataset}{https://www.kaggle.com/datasets/fedesoriano/hepatitis-c-dataset}) consists of 589 observations from a private hospital which are described by 14 predefined features. This dataset does not contain any missing values. It includes patients' liver function test results and doctors' observations for diagnosing and monitoring liver disease or damage.\\
\textit{Stroke Prediction}: The dataset (\href{https://www.kaggle.com/datasets/fedesoriano/stroke-prediction-dataset}{https://www.kaggle.com/datasets/fedesoriano/stroke-prediction-dataset}) consists of 110 observations which are described by 12 features. The aim is to predict the occurrence of stroke based on factors such as gender, age, various diseases, and smoking status.\\
\textit{Psychologist notes}: This dataset (\href{https://www.kaggle.com/datasets/shariful07/student-mental-health}{https://www.kaggle.com/datasets/shariful07/student-mental-health}) contains 51 records of psychologist's observations gathered from private US colleges. It comprises 11 descriptive elements and provides insight into students' mental health needs, along with the prospective benefits of additional consultations and their categorization. Due to its size this dataset is not used in the evaluation of predictive tasks, but only to assess the quality of extraction of tabular information from text. 

\textbf{Data privacy.}
The  Patient treatment classification and Heart disease diagnosis datasets are gathered in the form of medical reports from a single clinic. The data has been thoroughly anonymized to ensure privacy. Data collection and usage align with a strict privacy policy, adhering to all legal and ethical standards.

\textbf{Compared models.} We compare TEMED-LLM variants based on logistic regression ({\myfont logreg}), decision tree ({\myfont dtree}), and XGboost ({\myfont xgboost})) with established baselines such as BERT-base-cased ({\myfont bert-base-cased}) \citep{devlin2018bert}, RoBERTa-base ({\myfont roberta-base}) \citep{liu2019roberta}, XLM-RoBERTa-large ({\myfont xlm-roberta-large}) \citep{conneau2019unsupervised}, BioBERT ({\myfont dmis-lab/biobert-v1.1})\citep{lee2020biobert}, Set Fit ({\myfont set-fit-mpnetv2}) \citep{tunstall2022efficient}. All of these models are trained on the training set. We also include in the comparison InstructGPT ({\myfont text-davinchi-003}) which is used in a 10-shot learning regime (where shots come from the training set). The backbone LLM model in TEMED-LLM is InstructGPT ({\myfont text-davinchi-003}), whereas in the appendix~\ref{app:extract-rextract}, we also provided results for ChatGPT ({\myfont gpt3.5}) that are slightly worse. Additionally, details of model setups are provided in appendix~\ref{app:details}.

%\textbf{Experimental stages.} The experimental evaluation proceeds in three stages. First we compare prompt generation techniques for tabular data extraction. Subsequently, we assess the predictive performance of TEMED-LLM in comparison to that of baseline text classifiers. Finally, we evaluate the interpretability of the TEMED-LLM model.

\subsection{Evaluation of tabular data extraction}
\label{subsec:extreval}

We compare different prompt generation techniques for extracting predefined features from tabular data embedded in medical reports. These techniques are: {\myfont Extract-only}, {\myfont RExtract}, and {\myfont RExtract + VORC}. 
They are evaluated across all five datasets, and examples of {\myfont RExtract} prompts for each dataset are provided in appendix~\ref{app:prompts-rextract}. The evaluation is performed by comparing the ground truth tabular data with the tabular data extracted from medical reports. The comparison is performed primarily in terms of accuracy.  When computing the accuracy, the true positive instances are only those which perfectly match the corresponding ground truth instances. We also evaluate the successful detection of missing values in terms of precision and recall. These metrics will deteriorate if the model hallucinates values instead of missing ones or reports a missing value where the value is actually present. Moreover, we report an additional metric, {\myfont VORC} calls, which reveals the percentage of the total number of samples where {\myfont VORC} provided at least one feedback to the LLM.

The results, as displayed in Table~\ref{tab:Txt2tab}, reveal that {\myfont RExtract + VORC} offers the best performance metrics in all cases. {\myfont VORC} significantly enhances the accuracy of extraction and the detection of missing values. It shows impressive performance on the Hepatitis dataset (acc. $0.99$), and the highest difference between {\myfont RExtract} and {\myfont RExtract + VORC} (acc. difference $0.07$) is achieved on Heart disease diagnosis. However, {\myfont REextract+VORC} struggles on the Psychologist notes dataset. Upon examining the prompt examples (provided in the appendix), it can be concluded that in the cases in which the features sought in the text are explicitly mentioned, like in the Hepatitis prediction dataset, {\myfont RExtract+ VORC} extracts them successfully. However, in the cases in which the features are embedded in less structured text, like in the Psychologist notes dataset, the performance deteriorates. 

The average percentage of {\myfont VORC} calls is approximately $10\%$. This percentage tends to be higher in instances where REextract faces difficulties in extracting features from medical reports, such as in Psychologist notes ($14.20\%$) and Heart disease diagnosis ($12.00\%$), and lower in cases where feature extraction is straightforward, like Patient treatment classification ($2.54\%$).

\begin{table}[]
\centering
  \caption{Comparison of different extraction techniques, evaluated based on extraction accuracy, precision and recall with respect to missing values, and percentage of {\myfont VORC} calls.}
  \label{tab:Txt2tab}
\begin{tabular}{lcccc}
                & \multicolumn{4}{c}{\textbf{Heart disease diagnosis}}                                        \\ \cline{1-5} 
\textbf{model}  & \textbf{acc.} & \textbf{prec. (missing)} & \textbf{rec. (missing)} & \textbf{{\myfont VORC} calls} \\ \hline
{\myfont Extract-only}    & 0.57          & 0.37                 & 0.68                & -                   \\
{\myfont RExtract}        & 0.91          & 0.83                 & 0.92                 & -                   \\
{\myfont RExtract + VORC} & \textbf{0.98}          & \textbf{0.97}                 & \textbf{0.99}                & 12.00\%             \\ \hline
                & \multicolumn{4}{c}{\textbf{Patient treatment classification}}                      \\ \hline
{\myfont Extract-only}    & 0.89          & 0.82                  & 0.84                & -                   \\
{\myfont RExtract}        & 0.95          & 0.93                 & 0.94                & -                   \\
{\myfont RExtract + VORC} & \textbf{0.96}          & \textbf{0.97}                 & \textbf{0.96}                 & 2.54\%              \\ \hline
                & \multicolumn{4}{c}{\textbf{Hepatitis prediction}}                                  \\ \hline
{\myfont Extract-only}    & 0.88          & -                     & -                    & -                   \\
{\myfont RExtract}        & 0.92          & -                     & -                    & -                   \\
{\myfont RExtract + VORC} & \textbf{0.99}          & -                     & -                    & 8.90\%              \\ \hline
                & \multicolumn{4}{c}{\textbf{Stroke prediction}}                                     \\ \hline
{\myfont Extract-only}    & 0.73          & 0.52                 & 0.77                 & -                   \\
{\myfont RExtract}        & 0.87          & 0.86                 & 0.87                & -                   \\
{\myfont RExtract + VORC} & \textbf{0.92}          & \textbf{0.89}                 & \textbf{0.90}                & 11.00\%       \\ \hline
                & \multicolumn{4}{c}{\textbf{Psychologist notes}}                                    \\ \hline
{\myfont Extract-only}    & 0.71          & 0.75                 & \textbf{1.00}                & -                   \\
{\myfont RExtract}        & 0.78          & \textbf{1.00}                     & \textbf{1.00}                    & -                   \\
{\myfont RExtract + VORC} & \textbf{0.84}          & \textbf{1.00}                    & \textbf{1.00}                    & 14.20\%                  
\end{tabular}
\end{table}

\subsection{Evaluation of predictive performance}

After the tabular data has been extracted, we model that data by tabular ML methods. We compare the obtained tabular models with advanced text classification models. The objective is to demonstrate that tabular data models, trained on carefully extracted features from medical reports, can deliver superior predictive performance to even the most sophisticated text classifiers. This comparison is undertaken across four different datasets, using classification performance metrics such as accuracy, precision, recall, F1 score, and AUC.

In Table~\ref{tab:TxtvsTab} we presented the classification performance across all datasets. Results indicate that the {\myfont xgboost} model consistently outperformed other models across all datasets. Interestingly, the performance of interpretable tabular data models, namely {\myfont logreg} and {\myfont dtree}, is comparable to the {\myfont set-fit-mpnetv2} model. For instance, in the Heart disease diagnosis dataset, the AUC of {\myfont logreg} ($0.860$) and of {\myfont dtree} ($0.901$) was somewhat less than that of {\myfont set-fit-mpnetv2} ($0.931$), whereas, for the Hepatitis prediction dataset, their AUC was better. Furthermore, the text classification model trained on biomedical data, {\myfont dmis-lab/biobert-v1.1}, exhibited inferior performance compared to the interpretable tabular models. 

The greatest range of variation for any discussed metric on any of the datasets over five repeated experiments is $0.031$. The average range is $0.015$ and the smallest is $0.001$. Thus exhibited random variation does not affect our conclusions.

\begin{table}[]
\centering
\caption{Comparison of textual ML models applied on raw medical reports and tabular data models (TEMED-LLM) derived from extracted tabular data, across four distinct datasets.
} 
\label{tab:TxtvsTab}
\begin{adjustbox}{max width=\linewidth}
\begin{tabular}{lllllllllll}
                      & \multicolumn{5}{c}{\textbf{Heart disease diagnosis}}                                                                                                                                       & \multicolumn{5}{c}{\textbf{Patient treatment classification}}                                                                                                                   \\ \hline
\textbf{model}        & \multicolumn{1}{c}{\textbf{acc.}} & \multicolumn{1}{c}{\textbf{prec.}} & \multicolumn{1}{c}{\textbf{rec.}} & \multicolumn{1}{c}{\textbf{F1}} & \multicolumn{1}{c|}{\textbf{AUC}} & \multicolumn{1}{c}{\textbf{acc.}} & \multicolumn{1}{c}{\textbf{prec.}} & \multicolumn{1}{c}{\textbf{rec.}} & \multicolumn{1}{c}{\textbf{F1}} & \multicolumn{1}{c}{\textbf{AUC}} \\ \hline
{\myfont logreg}                & 0.860                             & 0.869                              & 0.906                             & 0.887                           & \multicolumn{1}{l|}{0.924}        & 0.643                             & 0.612                              & 0.448                             & 0.517                           & 0.682                            \\
{\myfont dtree}                 & 0.901                             & 0.917                              & 0.919                             & 0.918                           & \multicolumn{1}{l|}{0.947}        & 0.643                             & 0.585                              & 0.567                             & 0.576                           & 0.658                            \\
{\myfont xgboost}               & \textbf{0.985}                             & \textbf{0.984}                              & \textbf{0.991}                             & \textbf{0.987}                           & \multicolumn{1}{l|}{\textbf{0.998}}        & \textbf{0.669}                             & \textbf{0.619}                              & \textbf{0.582}                             & \textbf{0.600}                           & \textbf{0.719}                            \\ \hline
{\myfont xlm-roberta-large}     & 0.604                             & 0.302                              & 0.522                             & 0.382                           & \multicolumn{1}{l|}{0.855}        & 0.587                             & 0.321                              & 0.533                             & 0.401                           & 0.641                            \\
{\myfont roberta-base}          & 0.604                             & 0.302                              & 0.522                             & 0.382                           & \multicolumn{1}{l|}{0.758}        & 0.573                             & 0.287                              & 0.500                             & 0.364                           & 0.609                            \\
{\myfont bert-base-cased}       & 0.715                             & 0.722                              & 0.668                             & 0.671                           & \multicolumn{1}{l|}{0.737}        & 0.599                             & 0.583                              & 0.572                             & 0.568                           & 0.591                            \\
{\myfont dmis-lab/biobert-v1.1} & 0.708                             & 0.756                              & 0.644                             & 0.637                           & \multicolumn{1}{l|}{0.674}        & 0.618                             & 0.606                              & 0.602                             & 0.602                           & 0.614                            \\
{\myfont set-fit-mpnetv2}       & 0.931                             & 0.933                              & 0.954                             & 0.943                           & \multicolumn{1}{l|}{0.951}        & 0.631                             & 0.567                              & 0.567                             & 0.567                           & 0.671                            \\ \hline
{\myfont text-davinci-003}      & \multicolumn{1}{c}{0.620}         & \multicolumn{1}{c}{0.451}          & \multicolumn{1}{c}{0.535}         & \multicolumn{1}{c}{0.488}       & \multicolumn{1}{c|}{None}         & \multicolumn{1}{c}{0.598}         & \multicolumn{1}{c}{0.389}          & \multicolumn{1}{c}{0.551}         & \multicolumn{1}{c}{0.456}       & \multicolumn{1}{c}{None}         \\ \hline
                      & \multicolumn{5}{c|}{\textbf{Hepatitis prediction}}                                                                                                                               & \multicolumn{5}{c}{\textbf{Stroke prediction}}                                                                                                                                  \\ \hline
{\myfont logreg}                & 0.966                             & 0.889                              & 0.727                             & 0.800                           & \multicolumn{1}{l|}{0.946}        & 0.740                             & 0.707                              & 0.820                             & 0.759                           & 0.787                            \\
{\myfont dtree}                 & 0.983                             & 1.000                              & 0.818                             & 0.900                           & \multicolumn{1}{l|}{0.910}        & 0.700                             & 0.679                              & 0.760                             & 0.717                           & 0.779                            \\
{\myfont xgboost}               & \textbf{0.992}                             & \textbf{1.000}                              & \textbf{0.909}                             & \textbf{0.952}                           & \multicolumn{1}{l|}{\textbf{0.997}}        & \textbf{0.740}                             & \textbf{0.731}                              & \textbf{0.760}                             & \textbf{0.745}                           & \textbf{0.792}                            \\ \hline
{\myfont xlm-roberta-large}     & 0.552                             & 0.432                              & 0.549                             & 0.483                           & \multicolumn{1}{l|}{0.652}        & 0.575                             & 0.420                              & 0.508                             & 0.459                           & 0.652                            \\
{\myfont roberta-base}          & 0.542                             & 0.350                              & 0.545                             & 0.426                           & \multicolumn{1}{l|}{0.547}        & 0.566                             & 0.310                              & 0.522                             & 0.388                           & 0.547                            \\
{\myfont bert-base-cased}       & 0.510                             & 0.512                              & 0.510                             & 0.492                           & \multicolumn{1}{l|}{0.539}        & 0.599                             & 0.512                              & 0.510                             & 0.511                           & 0.539                            \\
{\myfont dmis-lab/biobert-v1.1} & 0.580                             & 0.610                              & 0.580                             & 0.550                           & \multicolumn{1}{l|}{0.652}        & 0.620                             & 0.684                              & 0.672                             & 0.678                           & 0.652                            \\
{\myfont set-fit-mpnetv2}       & 0.949                             & 0.778                              & 0.636                             & 0.700                           & \multicolumn{1}{l|}{0.844}        & 0.700                             & 0.692                              & 0.720                             & 0.706                           & 0.748                            \\ \hline
{\myfont text-davinci-003}      & \multicolumn{1}{c}{0.642}         & \multicolumn{1}{c}{0.421}          & \multicolumn{1}{c}{0.652}         & \multicolumn{1}{c}{0.512}       & \multicolumn{1}{c|}{-}         & \multicolumn{1}{c}{0.558}         & \multicolumn{1}{c}{0.289}          & \multicolumn{1}{c}{0.601}         & \multicolumn{1}{c}{0.390}       & \multicolumn{1}{c}{-}         \\ \hline
\end{tabular}
\end{adjustbox}
\end{table}

\subsection{Evaluation of interpretability}
\label{Sec:interpretability}

Our claim on interpretability relies on two assumptions. The first one is the interpretability of tabular machine learning models which are applied. It is widely accepted that some models (e.g., logistic regression and decision trees) are interpretable, although arguments can be made that interpretation of even the simplest models is not a trivial task. Some models are more complex (e.g., XGBoost), but still admit limited interpretability by different kinds of techniques. We adopt this general view of the interpretability of tabular models. We do not discuss this issue further, as such issues are widely discussed elsewhere~\cite{molnar2022}. 

Another assumption is that the possible errors in the data extraction process do not considerably affect our models. If they did, any interpretation would be less meaningful. We base our evaluation of this assumption on the comparison of the models trained on ground truth tabular data and the models based on the extracted tabular data. We have two expectations if our assumption is true. The first one is that such models have similar predictive performance, demonstrating a degree of stability with respect to the errors introduced in the data. We measure the discrepancy in predictive performance by the absolute difference of accuracies (acc.-d) of the models and by the absolute difference of AUCs (AUC-d) of the models (the lower the better). Our second expectation is that such models assign similar importance to different features, demonstrating that they make their decisions in a similar manner. We measure the discrepancy of feature importances using $\mathrm{R^2}$ score over feature importance vectors of two pairs of models -- one trained on ground truth data and one trained on extracted data (the greater the better).

%We present an in-depth comparison of models trained on tabular ground truth data and models trained on tabular data extracted from medical reports. Our goal is to demonstrate that both types of models offer similar explanations for their output predictions, thus demonstrating that the extraction process is faithful and does not hinder interepretability. The metrics we use to compare the models in this sense include absolute accuracy difference (acc.-d) between , absolute AUC difference (AUC-d), and $\mathrm{R^2}$ score. We calculate accuracy difference and AUC difference by finding the absolute difference between the corresponding metrics of the models trained on ground truth data and those trained on the extracted tabular data. Lower values for these metrics indicate a greater similarity between the two types of models.The $\mathrm{R^2}$ score, on the other hand, is obtained by comparing the feature importances of the models trained on the ground truth data and the extracted tabular data. A higher $\mathrm{R^2}$ score (closer to 1) signifies that the feature importances are more similar across the two models.

The comparison of used models is presented in Table~\ref{tab:comparison}. As can be seen, the models achieve very high values of evaluation metrics on two datasets. Logistic regression and XGBoost achieve almost perfect values on two of them.  Decision tree models seem to yield the least favorable results. For further insight, we have included visualizations of the decision tree models in the appendix~\ref{app:dtree-vis}.
The least favorable performance is obtained in the case of the Stroke prediction dataset due to the greater extraction error, as already established. This is the consequence of the larger disparity between the ground truth data and the extracted tabular data for this particular dataset, as already established in subsection~\ref{subsec:extreval}. Based on these results we can conclude that the quality of the interpretation relates to the quality of extraction in a predictable and expected way.

\begin{table}[]
\centering
\caption{Comparison of models trained on ground truth tables and models trained on extracted tables.} 
\label{tab:comparison}
\begin{adjustbox}{max width = \linewidth}
\begin{tabular}{llll|lll|lll|lll}
\textbf{}      & \multicolumn{3}{c|}{\textbf{\begin{tabular}[c]{@{}c@{}}Hearth disease\end{tabular}}} & \multicolumn{3}{c|}{\textbf{\begin{tabular}[c]{@{}c@{}}Patient treatment\end{tabular}}} & \multicolumn{3}{c|}{\textbf{\begin{tabular}[c]{@{}c@{}}Hepatitis\end{tabular}}} & \multicolumn{3}{c}{\textbf{\begin{tabular}[c]{@{}c@{}}Stroke\end{tabular}}} \\
\textbf{model} & \textbf{acc.-d}  &  \textbf{AUC-d} & \textbf{$\mathbf{R^2}$} & \textbf{acc.-d} & \textbf{AUC-d} & \textbf{$\mathbf{R^2}$}                       & \textbf{acc.-d}  & \textbf{AUC-d} & \textbf{$\mathbf{R^2}$}                  & \textbf{acc.-d}& \textbf{AUC-d}& \textbf{$\mathbf{R^2}$}\\ \hline
{\myfont logreg}         & 0.008       & 0.004          & 0.999        & 0.009                            & 0.002                               & 0.922                            & 0.003                        & 0.001                          & 0.996                         & 0.057                       & 0.039                         & 0.918                       \\
{\myfont dtree}          & 0.001       & 0.001          & 0.977        & 0.015                            & 0.012                              & 0.921                             & 0.003                         & 0.002                         & 0.941                         & 0.036                       & 0.018                         & 0.928                       \\
{\myfont xgboost}        & 0.003       & 0.000          & 0.982        & 0.011                            & 0.012                               & 0.925                            & 0.004                        & 0.006                          & 0.998                         & 0.055                       & 0.013                         & 0.939                      
\end{tabular}
\end{adjustbox}
\end{table}

\section{Conclusions}
\label{Sec:Conclusion}

We proposed and developed a method of accurate and interpretable medical diagnostics based on the conversion of medical text to tabular data and on modeling that data by interpretable and computationally efficient machine learning models. The conversion is performed by an LLM empowered by a novel data validation and reasoning correction feedback loop. We demonstrated improvements in accuracy over the state-of-the-art text classification models. We demonstrated the faithfulness of data extraction and evaluated its relationship with the interpretability of the obtained models.

\textbf{Impact.} The medical impact of TEMED-LLM can be substantial. First, it offers better accuracy in the classification of medical records. Second, it does so using interpretable models, a major requirement in medical applications. Third, it does so using models which are easily trained. Specifically, they require less computational power and less specialized practitioners than state-of-the-art deep learning models. Also, prompting an LLM via an online service and training the tabular model on a conventional CPU machine is less costly than acquiring and maintaining the hardware resources needed to train deep learning models. Fourth, the extraction of tabular data from medical records allows for storing and managing medical data in various data management systems. Fifth, it allows for the automation of testing of various medical hypotheses over the extracted data. We deem the TEMED-LLM system well-positioned to contribute to the advancements in medical diagnostics and research and to boost practitioners' trust in the automation of medical decision-making.

\textbf{Limitations.} TEMED-LLM requires domain-related input from a medical expert. Its effectiveness depends on experts' ability to identify relevant features in the reports at hand. The expert's error cannot be compensated by the downstream modules of TEMED-LLM.

We conclude that TEMED-LLM with its effectiveness, simplicity, and interpretability makes a substantial step towards intelligent and trustworthy automation in medical decision-making.
 
%We utilized a unique reasoning step with generated prompts to convert unstructured text into structured data tables, which we then processed using interpretable ML models such as decision trees and logistic regression. This end-to-end approach enabled interpretable predictions and outperformed state-of-the-art text classification models in medical diagnostics and prediction tasks.
%We demonstrated the potential of LLMs to improve both the performance and trustworthiness of ML models in medical applications. Our TEMED-LLM model, due to its simplicity and interpretability, fosters confidence in decision-makers when adopting ML models for medical diagnostics. In conclusion, our work with TEMED-LLM represents a substantial step forward in the realm of medical text analysis, and we believe it will pave the way for more interpretable and reliable ML applications in healthcare. Future research may extend our approach to other domains that require the extraction of structured data from unstructured text, as well as investigate the integration of domain-specific knowledge and expert-guided learning to further enhance the model's performance and robustness.

\bibliography{literature}

\clearpage
\begin{appendix}
\section{Appendix}
\label{sec:appendix}

\subsection{Detailed Description of Datasets}
\label{app:dataset}
In this section, we provided a detailed description of all five datasets:

\textbf{{Heart disease diagnosis:}} The dataset encompasses 917 Electronic Health Records (EHRs) from a private healthcare facility, characterized by a set of 11 predetermined attributes. The records encompass varied information including patients' historical data, results from clinical examinations, diagnostic tests, preoperative and postoperative care, patients' progress, and prescribed medications. The overarching goal of the dataset is to ascertain the occurrence of heart disease in patients, which is a binary attribute. The following are the input features of the dataset:

\begin{itemize}
\item \textbf{Age}: This is an integer value that denotes the age of the patient in years.
\item \textbf{Sex}: This is a categorical value representing the sex of the patient. It can be 'M' for male or 'F' for female.
\item \textbf{Chest Pain Type}: This is a categorical value indicating the type of chest pain experienced by the patient. It can be 'TA' for Typical Angina, 'ATA' for Atypical Angina, 'NAP' for Non-Anginal Pain, or 'ASY' for Asymptomatic.
\item \textbf{Resting Blood Pressure (Resting Bp)}: This is an integer value representing the patient's resting blood pressure measured in millimeters of mercury (mm Hg).
\item \textbf{Cholesterol}: This integer value denotes the serum cholesterol level of the patient in milligrams per deciliter (mm/dl).
\item \textbf{Fasting Blood Sugar (Fasting Bs)}: This is a binary value (0 or 1), where 1 indicates that the patient's fasting blood sugar is greater than 120 mg/dl, and 0 otherwise.
\item \textbf{Resting Electrocardiogram (Resting Ecg)}: This categorical value describes the results of the patient's resting electrocardiogram. It can be 'Normal', 'ST' for having ST-T wave abnormality, or 'LVH' for showing probable or definite left ventricular hypertrophy by Estes' criteria.
\item \textbf{Maximum Heart Rate (Max Hr)}: This is an integer value representing the maximum heart rate achieved by the patient. It ranges from 60 to 202.
\item \textbf{Exercise-induced Angina (Exercise Angina)}: This is a categorical value (Y or N) indicating whether the patient experienced angina (chest pain) caused by exercise. 'Y' stands for yes and 'N' for no.
\item \textbf{Oldpeak}: This numeric value represents ST depression induced by exercise relative to rest.
\item \textbf{ST Slope (St Slope)}: This is a categorical value denoting the slope of the peak exercise ST segment. It can be 'Up' for upsloping, 'Flat' for flat, or 'Down' for downsloping.
\end{itemize}

\textbf{Patient treatment classification:} This dataset incorporates 718 Electronic Health Records (EHRs) obtained from a private healthcare institution, each of which is characterized by 10 predefined features. The primary aim of this dataset is to aid in the determination of a patient's treatment trajectory, categorizing them as either in-care or out-care. This decision is based on the laboratory test results and physicians' observations. The pre-specified attributes embedded in these records are:

\begin{itemize}
\item \textbf{Haematocrit}: This numerical attribute represents the patient's laboratory test result for haematocrit.
\item \textbf{Haemoglobin}: This numerical attribute denotes the patient's laboratory test result for haemoglobin.
\item \textbf{Erythrocyte}: This numerical attribute pertains to the patient's laboratory test result for erythrocyte count.
\item \textbf{Leucocyte}: This numerical attribute corresponds to the patient's laboratory test result for leucocyte count.
\item \textbf{Thrombocyte}: This numerical attribute signifies the patient's laboratory test result for thrombocyte count.
\item \textbf{MCH}: This numerical attribute stands for the patient's laboratory test result for Mean Corpuscular Haemoglobin (MCH).
\item \textbf{MCHC}: This numerical attribute represents the patient's laboratory test result for Mean Corpuscular Haemoglobin Concentration (MCHC).
\item \textbf{MCV}: This numerical attribute delineates the patient's laboratory test result for Mean Corpuscular Volume (MCV).
\item \textbf{Age}: This integer attribute denotes the patient's age.
\item \textbf{Sex}: This categorical attribute indicates the patient's gender. 'M' denotes male, and 'F' signifies female.
\end{itemize}

\textbf{{Hepatitis prediction:}} This dataset encompasses 589 records, each of which is characterized by 12 predetermined features. The data are publicly available from \href{https://www.kaggle.com/datasets/fedesoriano/hepatitis-c-dataset}{https://www.kaggle.com/datasets/fedesoriano/hepatitis-c-dataset} and do not exhibit any missing values. The dataset amalgamates patients' liver function test results and physicians' observations, thereby aiding in the diagnosis and surveillance of liver diseases or damage. The pre-specified attributes embedded in these records are:

\begin{itemize}
\item \textbf{Age}: This integer attribute signifies the age of the patient in years.
\item \textbf{Sex}: This categorical attribute represents the patient's sex. 'f' denotes female, and 'm' stands for male.
\item \textbf{ALB}: This numerical attribute corresponds to the amount of albumin in the patient's blood.
\item \textbf{ALP}: This numerical attribute signifies the amount of alkaline phosphatase in the patient's blood.
\item \textbf{ALT}: This numerical attribute pertains to the amount of alanine transaminase in the patient's blood.
\item \textbf{AST}: This numerical attribute delineates the amount of aspartate aminotransferase in the patient's blood.
\item \textbf{BIL}: This numerical attribute represents the amount of bilirubin in the patient's blood.
\item \textbf{CHE}: This numerical attribute corresponds to the amount of cholinesterase in the patient's blood.
\item \textbf{CHOL}: This numerical attribute stands for the amount of cholesterol in the patient's blood.
\item \textbf{CREA}: This numerical attribute depicts the amount of creatine in the patient's blood.
\item \textbf{GGT}: This numerical attribute signifies the amount of gamma-glutamyl transferase in the patient's blood.
\item \textbf{PROT}: This numerical attribute corresponds to the amount of protein in the patient's blood.
\end{itemize}

\textbf{{Stroke prediction:}} This dataset comprises 110 entries, each one characterized by 10 distinctive attributes. The dataset, publicly available at \href{https://www.kaggle.com/datasets/fedesoriano/stroke-prediction-dataset}{https://www.kaggle.com/datasets/fedesoriano/stroke-prediction-dataset}, intends to aid in the prediction of stroke occurrence, based on factors such as gender, age, prevalent diseases, and smoking status. The predefined features in these records include:

\begin{itemize}
\item \textbf{Choose Your Gender}: Categorical attribute denoting the gender of the individual. The options are 'Female' and 'Male'.
\item \textbf{Age}: A numerical attribute signifying the age of the individual.
\item \textbf{What Is Your Course}: Categorical attribute indicating the course the individual is enrolled in.
\item \textbf{Your Current Year Of Study}: Categorical attribute representing the current year of study. Options are 'year 1', 'year 2', 'year 3', and 'year 4'.
\item \textbf{What Is Your Cgpa}: Categorical attribute reflecting the current CGPA range of the individual. Options range from '0 - 1.99', '2.00 - 2.49', '2.50 - 2.99', to '3.00 - 3.49', and '3.50 - 4.00'.
\item \textbf{Marital Status}: Categorical attribute indicating the marital status of the individual. The options are 'yes' for married and 'no' for not married.
\item \textbf{Do You Have Depression}: Categorical attribute signifying whether the individual has depression. The options are 'yes' or 'no'.
\item \textbf{Do You Have Anxiety}: Categorical attribute showing whether the individual has anxiety. The options are 'yes' or 'no'.
\item \textbf{Do You Have Panic Attack}: Categorical attribute depicting whether the individual experiences panic attacks. The options are 'yes' or 'no'.
\item \textbf{Did You Seek Any Specialist For Treatment}: Categorical attribute indicating whether the individual has sought treatment from a specialist. The options are 'yes' or 'no'.
\end{itemize}

\textbf{Psychologist Notes} dataset, publicly available from \href{https://www.kaggle.com/datasets/shariful07/student-mental-health}{https://www.kaggle.com/datasets/shariful07/student-mental-health}, contains a collection of 51 records documenting the observations of psychologists from private colleges in the United States. The data comprises 10 key features that provide insight into the mental health needs of students, with information pertaining to the potential benefits of additional consultations and their respective categorization. This dataset is not used for the evaluation of predictive tasks due to its size, but rather for the assessment of the quality of extraction of tabular data from text. The predefined features in these records include:

\begin{itemize}
    \item \textbf{Choose Your Gender}: The gender of the student, denoted as either 'Female' or 'Male'.
    \item \textbf{Age}: The age of the student.
    \item \textbf{What Is Your Course}: The specific course in which the student is enrolled.
    \item \textbf{Your Current Year Of Study}: The current year of study of the student, categorized into 'year 1', 'year 2', 'year 3', or 'year 4'.
    \item \textbf{What Is Your CGPA}: The current CGPA range of the student, segmented into the following ranges: '0 - 1.99', '2.00 - 2.49', '2.50 - 2.99', '3.00 - 3.49', and '3.50 - 4.00'.
    \item \textbf{Marital Status}: The marital status of the student, marked as either 'yes' or 'no'.
    \item \textbf{Do You Have Depression}: A flag indicating whether the student has depression, noted as either 'yes' or 'no'.
    \item \textbf{Do You Have Anxiety}: A flag indicating whether the student experiences anxiety, noted as either 'yes' or 'no'.
    \item \textbf{Do You Have Panic Attack}: A flag indicating whether the student experiences panic attacks, denoted as either 'yes' or 'no'.
    \item \textbf{Did You Seek Any Specialist For Treatment}: An indicator showing whether the student has sought treatment from a specialist, denoted as either 'yes' or 'no'.
\end{itemize}

\subsection{
Evaluating {\myfont Extract-only}, {\myfont RExtract}, and {\myfont RExtract+ VORC} Techniques Using ChatGPT in the TEMED-LLM Framework}
\label{app:extract-rextract}

In this section, we carry out an additional comparison of three distinct extraction techniques: {\myfont Extract-only}, {\myfont RExtract}, and {\myfont RExtract+ VORC}. In contrast to the results presented in section~\ref{Sec:exp-evaluation}, where the LLM was {\myfont text-davinci-003}, in this case, we utilized the ChatGPT ({\myfont{gpt3.5}}) model. As before, the evaluation is based on extraction accuracy, precision and recall with respect to missing values, and the percentage of {\myfont VORC} calls. The analysis is conducted over all five datasets. Results are presented in the table~\ref{tab:app_Txt2tab}.

It can be observed that {\myfont text-davinci-003} obtained slightly better performance compared to the {\myfont{gpt3.5}}. The disparity in performance is reasonable considering {\myfont text-davinci-003} is specifically trained to follow instructions, which gives it an edge in these tasks. The performance in this evaluation parallels the one described in section~\ref{subsec:extreval}, so we conclude that the difference is not significant. Therefore, both models can be effectively employed in conjunction with the {\myfont RExtract+ VORC} methodology for extracting features from medical reports and texts.

\begin{table}[]
\centering
  \caption{Comparison of {\myfont Extract-only}, {\myfont RExtract}, and {\myfont RExtract+ VORC} extraction techniques across five datasets, evaluated based on extraction accuracy, precision and recall with respect to missing values, and percentage of VORC calls. The used LLM is {\myfont gpt3.5}.}
  \label{tab:app_Txt2tab}
\begin{tabular}{l|cccc}
                & \multicolumn{4}{c}{\textbf{Heart disease diagnosis}}                                        \\ \cline{2-5} 
\textbf{model}  & \textbf{acc.} & \textbf{prec. (missing)} & \textbf{rec. (missing)} & \textbf{VORC calls} \\ \hline
{\myfont Extract-only}    & 0.55          & 0.4                 & 0.61                & -                   \\
{\myfont RExtract}        & 0.83          & 0.72                 & 0.80                 & -                   \\
{\myfont RExtract + VORC} & \textbf{0.95}          & \textbf{0.92}                 & \textbf{0.99}                & 16.00\%             \\ \hline
                & \multicolumn{4}{c}{\textbf{Patient treatment classification}}                      \\ \hline
{\myfont Extract-only}    & 0.90          & 0.78                  & 0.87                & -                   \\
{\myfont RExtract}        & 0.94          & 0.91                 & 0.94                & -                   \\
{\myfont RExtract + VORC} & \textbf{0.94}          & \textbf{0.92}                 & \textbf{0.92}                 & 1.2\%              \\ \hline
                & \multicolumn{4}{c}{\textbf{Hepatitis prediction}}                                  \\ \hline
{\myfont Extract-only}    & 0.85          & -                     & -                    & -                   \\
{\myfont RExtract}        & 0.90          & -                     & -                    & -                   \\
{\myfont RExtract + VORC} & \textbf{0.97}          & -                     & -                    & 9.70\%              \\ \hline
                & \multicolumn{4}{c}{\textbf{Stroke prediction}}                                     \\ \hline
{\myfont Extract-only}    & 0.73          & 0.52                 & 0.77                 & -                   \\
{\myfont RExtract}        & 0.82          & 0.68                 & 0.92                & -                   \\
{\myfont RExtract + VORC} & \textbf{0.89}          & \textbf{0.88}                 & \textbf{0.89}                & 9.5\%            
\\ \hline
                & \multicolumn{4}{c}{\textbf{Psychologist notes}}                                    \\ \hline
{\myfont Extract-only}    & 0.69          & 0.50                 & \textbf{1.00}                & -                   \\
{\myfont RExtract}        & 0.72          & \textbf{0.75}                     & \textbf{1.00}                    & -                   \\
{\myfont RExtract + VORC} & \textbf{0.79}          & \textbf{1.00}                    & \textbf{1.00}                    & 8.45\%             
\end{tabular}
\end{table}

\subsection{Comparative Analysis of Feature Importance in Models Trained on Ground Truth Data Versus Tabular Data Extracted from Medical Reports}
\label{app:feature-importance}

In this section, we provide a detailed comparison of feature importance for three tabular ML models: logistic regression ({\myfont logreg}), decision tree ({\myfont dtree}), and XGBoost (\myfont xgboost). The feature importance scores for {\myfont xgboost} and {\myfont dtree} models are normalized so as to sum to $1$. The importance of a feature in {\myfont dtree} is computed as the (normalized) total reduction of the criterion brought by that feature. For the {\myfont logreg} model for feature importance scores, we take the corresponding weights of the model.
We present the comparison in Table~\ref{tab:my_label_1} and Table~\ref{tab:my_label_2}. Table~\ref{tab:my_label_1} presents the comparison on two datasets: Hepatitis prediction and Patient treatment classification. Table~\ref{tab:my_label_2} presents the comparison on Stroke prediction and Heart disease datasets. We present the importance of every feature of each dataset for each model. Moreover, we consider two kinds of models -- models trained on tabular data extracted from medical reports and models trained on ground truth tabular data. The models trained on ground truth tabular data are denoted with~{\myfont \_gt}. It is these importance scores that are used to evaluate the $R^2$ values presented in table~\ref{tab:comparison}.

Upon examining the evaluation data, it is evident that the differences in feature importance of models trained on extracted and ground truth tabular data, across all datasets are negligible. 

\begin{table}[]
    \centering
    \caption{Comparison of feature importance scores for models trained on Hepatitis prediction and Patient treatment classification datasets. We compare the models trained on tabular data extracted from  medical reports  and the models trained on ground truth tabular data.}
    
\begin{tabular}{lllllll}
\hline
\multicolumn{1}{l|}{}                 & {\myfont xgboost} & {\myfont xgboost\_gt} & {\myfont dtree}  & {\myfont dtree\_gt} & \multicolumn{1}{l}{\myfont logreg} & {\myfont logreg\_gt} \\ \hline \hline
\multicolumn{7}{l}{\textbf{Hepatitis prediction}}                                                                                      \\ \hline \hline
\multicolumn{1}{l|}{Age}              & 0.004   & 0.003       & 0.015  & 0.000     & 0.002                       & -0.090     \\ \hline
\multicolumn{1}{l|}{Laboratory\_ALB}  & 0.030   & 0.030       & 0.004  & 0.000     & -0.495                      & -0.420     \\ \hline
\multicolumn{1}{l|}{Laboratory\_ALP}  & 0.083   & 0.084       & 0.022  & 0.030     & -1.537                      & -1.526     \\ \hline
\multicolumn{1}{l|}{Laboratory\_ALT}  & 0.167   & 0.167       & 0.240  & 0.255     & -2.651                      & -2.576     \\ \hline
\multicolumn{1}{l|}{Laboratory\_AST}  & 0.531   & 0.531       & 0.580  & 0.607     & 1.844                       & 1.882      \\ \hline
\multicolumn{1}{l|}{Laboratory\_BIL}  & 0.077   & 0.077       & 0.002  & 0.000     & 0.862                       & 1.066      \\ \hline
\multicolumn{1}{l|}{Laboratory\_CHE}  & 0.023   & 0.023       & 0.000  & 0.036     & 0.425                       & 0.337      \\ \hline
\multicolumn{1}{l|}{Laboratory\_CHOL} & 0.012   & 0.012       & 0.001  & 0.000     & -0.666                      & -0.565     \\ \hline
\multicolumn{1}{l|}{Laboratory\_CREA} & 0.002   & 0.003       & 0.030  & 0.020     & 0.903                       & 0.917      \\ \hline
\multicolumn{1}{l|}{Laboratory\_GGT}  & 0.072   & 0.062       & 0.039  & 0.052     & 1.705                       & 1.695      \\ \hline
\multicolumn{1}{l|}{Laboratory\_PROT} & 0.007   & 0.007       & 0.120  & 0.000     & 0.694                       & 0.753      \\ \hline
\multicolumn{1}{l|}{Sex\_female}      & 0.000   & 0.000       & -0.065 & 0.000     & 0.222                       & 0.207      \\ \hline
\multicolumn{1}{l|}{Sex\_male}        & 0.000   & 0.000       & 0.000  & 0.000     & -0.241                      & -0.206     \\ \hline \hline
\multicolumn{7}{l}{\textbf{Patient treatment classification}}                                                                          \\ \hline \hline
\multicolumn{1}{l|}{Haematocrit}      & 0.101   & 0.105       & 0.001  & 0.000     & -0.164                      & -0.219     \\ \hline
\multicolumn{1}{l|}{Haemoglobins}     & 0.127   & 0.166       & 0.285  & 0.330     & -0.349                      & -0.271     \\ \hline
\multicolumn{1}{l|}{Erythrocyte}      & 0.100   & 0.076       & 0.061  & 0.096     & -0.426                      & -0.227     \\ \hline
\multicolumn{1}{l|}{Leucocyte}        & 0.094   & 0.090       & 0.142  & 0.156     & 0.211                       & 0.280      \\ \hline
\multicolumn{1}{l|}{T{]}hrombocyte}   & 0.120   & 0.129       & 0.239  & 0.261     & -0.555                      & -0.505     \\ \hline
\multicolumn{1}{l|}{MCH}              & 0.082   & 0.092       & 0.003  & 0.023     & -0.341                      & -0.284     \\ \hline
\multicolumn{1}{l|}{MCHC}             & 0.083   & 0.099       & 0.054  & 0.045     & -0.009                      & 0.057      \\ \hline
\multicolumn{1}{l|}{MCV}              & 0.070   & 0.064       & 0.064  & 0.000     & -0.021                      & 0.027      \\ \hline
\multicolumn{1}{l|}{Age}              & 0.108   & 0.103       & 0.089  & 0.088     & 0.021                       & 0.089      \\ \hline
\multicolumn{1}{l|}{Sex\_female}      & 0.065   & 0.076       & 0.002  & 0.000     & -0.276                      & -0.246     \\ \hline
\multicolumn{1}{l|}{Sex\_male}        & 0.026   & 0.000       & 0.062  & 0.000     & 0.117                       & 0.246      \\ \hline
\end{tabular}

    \label{tab:my_label_1}
\end{table}

\begin{table}[]
    \centering
    \caption{
    Comparison of feature importance scores for models trained on Stroke prediction and Heart disease diagnosis datasets. We compare the models trained on tabular data extracted from  medical reports  and the models trained on ground truth tabular data.}
    \begin{tabular}{lllllll}
\hline
\multicolumn{1}{l|}{}                                 & {\myfont xgboost} & {\myfont xgboost\_gt} & {\myfont dtree} & {\myfont dtree\_gt} & {\myfont logreg} & {\myfont logreg\_gt} \\ \hline \hline
\multicolumn{7}{l}{\textbf{Stroke prediction}}                                                                                   \\ \hline \hline
\multicolumn{1}{l|}{Age}                              & 0.197   & 0.186       & 0.601 & 0.751     & 1.728  & 1.735      \\ \hline
\multicolumn{1}{l|}{Avg\_glucose\_level}              & 0.044   & 0.049       & 0.000 & 0.044     & 0.163  & 0.131      \\ \hline
\multicolumn{1}{l|}{BMI}                              & 0.075   & 0.061       & 0.086 & 0.147     & 0.130  & 0.144      \\ \hline
\multicolumn{1}{l|}{Gender\_female}                   & 0.062   & 0.062       & 0.000 & 0.000     & -0.012 & 0.024      \\ \hline
\multicolumn{1}{l|}{Gender\_male}                     & 0.007   & 0.000       & 0.013 & 0.013     & 0.046  & -0.024     \\ \hline
\multicolumn{1}{l|}{Ever\_married\_No}                & 0.056   & 0.055       & 0.000 & 0.000     & 0.113  & 0.168      \\ \hline
\multicolumn{1}{l|}{Ever\_married\_Yes}               & 0.010   & 0.000       & 0.000 & 0.001     & -0.153 & -0.168     \\ \hline
\multicolumn{1}{l|}{Work\_type\_Govt\_job}            & 0.042   & 0.035       & 0.017 & 0.000     & -0.845 & -0.270     \\ \hline
\multicolumn{1}{l|}{Work\_type\_Never\_worked}        & 0.008   & 0.000       & 0.022 & 0.000     & -0.121 & -0.027     \\ \hline
\multicolumn{1}{l|}{Work\_type\_Private}              & 0.066   & 0.066       & 0.008 & 0.000     & -0.027 & -0.072     \\ \hline
\multicolumn{1}{l|}{Work\_type\_Self-employed}        & 0.028   & 0.051       & 0.000 & 0.025     & -0.232 & -0.323     \\ \hline
\multicolumn{1}{l|}{Work\_type\_children}             & 0.015   & 0.000       & 0.000 & 0.000     & 0.706  & 0.692      \\ \hline
\multicolumn{1}{l|}{Residence\_type\_Rural}           & 0.053   & 0.056       & 0.004 & 0.000     & 0.013  & -0.037     \\ \hline
\multicolumn{1}{l|}{Residence\_type\_Urban}           & 0.000   & 0.000       & 0.043 & 0.000     & -0.083 & 0.036      \\ \hline
\multicolumn{1}{l|}{Smoking\_status\_Unknown}         & 0.045   & 0.043       & 0.032 & 0.000     & 0.121  & 0.160      \\ \hline
\multicolumn{1}{l|}{Smoking\_status\_formerly smoked} & 0.056   & 0.048       & 0.000 & 0.000     & -0.069 & -0.037     \\ \hline
\multicolumn{1}{l|}{Smoking\_status\_never smoked}    & 0.047   & 0.066       & 0.035 & 0.000     & -0.178 & -0.169     \\ \hline
\multicolumn{1}{l|}{Smoking\_status\_smokes}          & 0.069   & 0.080       & 0.000 & 0.005     & -0.079 & 0.046      \\ \hline
\multicolumn{1}{l|}{Hypertension\_0}                  & 0.076   & 0.058       & 0.000 & 0.013     & -0.260 & -0.257     \\ \hline
\multicolumn{1}{l|}{Hypertension\_1}                  & 0.008   & 0.000       & 0.000 & 0.000     & 0.267  & 0.256      \\ \hline
\multicolumn{1}{l|}{Heart\_disease\_0}                & 0.092   & 0.083       & 0.000 & 0.000     & -0.072 & -0.073     \\ \hline
\multicolumn{1}{l|}{Heart\_disease\_1}                & 0.010   & 0.000       & 0.140 & 0.000     & 0.121  & 0.073      \\ \hline \hline
\multicolumn{7}{l}{\textbf{Heart disease diagnosis}}                                                                             \\ \hline \hline
\multicolumn{1}{l|}{Age}                              & 0.013   & 0.015       & 0.028 & 0.028     & 0.122  & 0.129      \\ \hline
\multicolumn{1}{l|}{RestingBP}                        & 0.014   & 0.013       & 0.025 & 0.025     & 0.099  & 0.066      \\ \hline
\multicolumn{1}{l|}{Cholesterol}                      & 0.014   & 0.013       & 0.000 & 0.045     & -0.514 & -0.504     \\ \hline
\multicolumn{1}{l|}{MaxHR}                            & 0.088   & 0.019       & 0.139 & 0.138     & -0.258 & -0.249     \\ \hline
\multicolumn{1}{l|}{Oldpeak}                          & 0.022   & 0.020       & 0.087 & 0.095     & 0.458  & 0.460      \\ \hline
\multicolumn{1}{l|}{Sex\_female}                      & 0.036   & 0.035       & 0.028 & 0.028     & -0.658 & -0.656     \\ \hline
\multicolumn{1}{l|}{Sex\_male}                        & 0.000   & 0.000       & 0.000 & 0.000     & 0.657  & 0.656      \\ \hline
\multicolumn{1}{l|}{ChestPainType\_ASY}               & 0.079   & 0.082       & 0.159 & 0.158     & 1.212  & 1.215      \\ \hline
\multicolumn{1}{l|}{ChestPainType\_ATA}               & 0.016   & 0.002       & 0.000 & 0.000     & -0.356 & -0.379     \\ \hline
\multicolumn{1}{l|}{ChestPainType\_NAP}               & 0.016   & 0.016       & 0.000 & 0.000     & -0.533 & -0.523     \\ \hline
\multicolumn{1}{l|}{ChestPainType\_TA}                & 0.002   & 0.003       & 0.000 & 0.000     & -0.323 & -0.314     \\ \hline
\multicolumn{1}{l|}{FastingBS\_0}                     & 0.037   & 0.030       & 0.021 & 0.025     & -0.485 & -0.489     \\ \hline
\multicolumn{1}{l|}{FastingBS\_1}                     & 0.000   & 0.000       & 0.005 & 0.000     & 0.485  & 0.489      \\ \hline
\multicolumn{1}{l|}{RestingECG\_LVH}                  & 0.016   & 0.014       & 0.000 & 0.000     & 0.149  & 0.143      \\ \hline
\multicolumn{1}{l|}{RestingECG\_Normal}               & 0.000   & 0.006       & 0.000 & 0.000     & 0.030  & 0.025      \\ \hline
\multicolumn{1}{l|}{RestingECG\_ST}                   & 0.009   & 0.015       & 0.000 & 0.000     & -0.179 & -0.168     \\ \hline
\multicolumn{1}{l|}{ExerciseAngina\_N}                & 0.000   & 0.027       & 0.024 & 0.024     & -0.505 & -0.507     \\ \hline
\multicolumn{1}{l|}{ExerciseAngina\_Y}                & 0.000   & 0.000       & 0.000 & 0.000     & 0.505  & 0.507      \\ \hline
\multicolumn{1}{l|}{ST\_Slope\_Down}                  & 0.017   & 0.024       & 0.000 & 0.000     & -0.224 & -0.224     \\ \hline
\multicolumn{1}{l|}{ST\_Slope\_Flat}                  & 0.019   & 0.017       & 0.049 & 0.000     & 1.053  & 1.059      \\ \hline
\multicolumn{1}{l|}{ST\_Slope\_Up}                    & 0.601   & 0.649       & 0.435 & 0.435     & -0.828 & -0.835     \\ \hline
\end{tabular}

    \label{tab:my_label_2}
\end{table}

\subsection{Showcasing Examples of Prompts Using the {\myfont RExtract} Module}
\label{app:prompts-rextract}

In this section, we present some exemplary prompts for the {\myfont RExtract} module for all five datasets under consideration: Heart disease diagnosis, Patient treatment classification, Hepatitis prediction, Stroke prediction, and Psychological notes. Each prompt is separated into three sections: instructions (with one-shot examples), reasoning, and the medical text of interest. 

\subsubsection{Hearth disease diagnosis}

\begin{tcolorbox}[width=\textwidth, title =  {Hearth disease},
    enhanced, 
    size=small,
    skin first=enhanced,
    skin middle=enhanced,
    skin last=enhanced,
    ]{}
    {\myfont
    \colorbox{yellow}{\textbf{Instructions:}}The output JSON should be formatted as a JSON instance that conforms to the JSON schema from Pydantic.\\

As an example, for the schema \{"properties": \{"foo": \{"title": "Foo", "description": "a list of strings", "type": "array", "items": \{"type": "string"\}\}\}, "required": ["foo"]\}\}
the object \{"foo": ["bar", "baz"]\} is a well-formatted instance of the schema. The object \{"properties": \{"foo": ["bar", "baz"]\}\} is not well-formatted.\\

\colorbox{yellow}{\textbf{Here is the output JSON schema:}}\\
```
\{"properties": \{"age": \{"title": "Age", "description": "Age of the patient [int](years)", "type": "integer"\}, "sex": \{"title": "Sex", "description": "Sex of the patient [M,F] where M: Male, F: Female", "type": "string"\}, "chest\_pain\_type": \{"title": "Chest Pain Type", "description": "Chest pain type [ATA, NAP, ASY, TA] where TA: Typical Angina, ATA: Atypical Angina, NAP: Non-Anginal Pain, ASY: Asymptomatic", "type": "string"\}, "resting\_bp": \{"title": "Resting Bp", "description": "Resting blood pressure [int](mm Hg)", "type": "integer"\}, "cholesterol": \{"title": "Cholesterol", "description": "Serum cholesterol [int[(mm/dl)", "type": "integer"\}, "fasting\_bs": \{"title": "Fasting Bs", "description": "Fasting blood sugar [1,0] where 1: if FastingBS > 120 mg/dl, 0: otherwise", "type": "integer"\}, "resting\_ecg": \{"title": "Resting Ecg", "description": "Resting electrocardiogram results [Normal, ST, LVH] where Normal: Normal, ST: having ST-T wave abnormality (T wave inversions and/or ST elevation or depression of > 0.05 mV), LVH: showing probable or definite left ventricular hypertrophy by Estes' criteria", "type": "string"\}, "max\_hr": \{"title": "Max Hr", "description": "Maximum heart rate achieved [Numeric value between 60 and 202]", "type": "integer"\}, "exercise\_angina": \{"title": "Exercise Angina", "description": "Exercise-induced angina [Y,N] where Y: Yes, N: No", "type": "string"\}, "oldpeak": \{"title": "Oldpeak", "description": "Oldpeak = ST Numeric value measured in depression", "anyOf": [\{"type": "number"\}, \{"type": "integer"\}]\}, "st\_slope": \{"title": "St Slope", "description": "The slope of the peak exercise ST segment [Up, Flat, Down] where Up: upsloping, Flat: flat, Down: downsloping", "type": "string"\}\}\}
```\\

\colorbox{yellow}{\textbf{When generating JSON instance follow this format:}}

Medical report: the input medical report from which you should extract JSON instance.\\
Reasoning: give me an explanation of how you assign value for a given key. Thinking step by step for each key before assigning a value to it.\\
Output JSON: The final output JSON should be formatted as a JSON instance that conforms to the output JSON schema above.\\

\colorbox{yellow}{\textbf{Here is an example of a process:}}\\
Medical report:\\
"REASON FOR CONSULTATION: Chest pain.\\
HISTORY OF PRESENT ILLNESS: The patient is a 63-year-old Caucasian male with a past medical history significant for hypertension and hyperlipidemia. He presents with chest pain that occurs during moderate physical activity. The patient has a history of smoking and occasional alcohol consumption.\\
PAST MEDICAL HISTORY: Hypertension. Hyperlipidemia. Type 2 Diabetes.\\
ALLERGIES: No known drug allergies.\\

}
\end{tcolorbox}

\begin{tcolorbox}[width=\textwidth, notitle,
    size=small,
    ]{}
    {\myfont FAMILY HISTORY: Father had a myocardial infarction at the age of 65.\\
SOCIAL HISTORY: 20-pack-year smoking history, occasional alcohol consumption, and no illicit drug use.\\
CURRENT MEDICATIONS: Amlodipine, Atorvastatin, Metformin, Aspirin.\\
REVIEW OF SYSTEMS: Patient reports occasional shortness of breath and dizziness.\\
PHYSICAL EXAMINATION:\\
VITAL SIGNS: Blood pressure 145/90, pulse 100, O2 saturation 98\% on room air, temperature 98.6, respiratory rate 18.\\
GENERAL: Patient is alert and oriented, and appears comfortable at rest.\\
HEAD AND NECK: No JVP has been seen, no carotid bruits.\\
CHEST: Clear to auscultation bilaterally.\\
CARDIOVASCULAR: Regular rhythm, standard S1, and S2, no murmurs, rubs, or gallops.\\
ABDOMEN: Soft, non-tender, and non-distended.\\
EXTREMITIES: No edema or clubbing.\\
DATA: A 12-lead EKG revealed normal sinus rhythm with a rate of 100 beats per minute and nonspecific ST-T wave abnormality.\\
LABORATORY DATA: WBC 6.8, hemoglobin 14.2, platelets 250. Chemistry-7 within normal limits. Lipid profile: Triglycerides 160, total cholesterol 220, HDL 40, LDL 140. Fasting blood sugar 125 mg/dL. Liver function tests within normal limits. BUN and creatinine are within normal limits.\\
TREADMILL STRESS TEST: The patient achieved a maximum heart rate of one hundred and fifty beats per minute. No chest pain or shortness of breath during the test. An old peak of 1.5 mm was noted in leads II, III, and VF. Shift relative to exercise-induced increments in heart rate is showing a downward trend.\\

\colorbox{orange}{\textbf{Reasoning:}}\\
The report states that the patient is a 63-year-old Caucasian male, therefore "Age": 63.\\
The patient is described as a male in the report, therefore "Sex": "M".\\
The patient described experiencing chest discomfort without typical symptoms, such as shortness of breath, dizziness, or sweating. Based on this information, the most appropriate 'ChestPainType' for this patient would be "ASY", which stands for "asymptomatic." Therefore ChestPainType: ASY.\\
The patient's blood pressure is mentioned in the "VITAL SIGNS" section as 145/90, hence "RestingBP": 145.\\
The report mentioned the patient's total cholesterol value under "LABORATORY DATA": "total cholesterol 220", therefore "Cholesterol": 220.\\
The fasting blood sugar is mentioned in the "LABORATORY DATA" section as 125 mg/dL, which is greater than 120 mg/dL, so the value is 1, therefore "FastingBS": 1.\\
The 12-lead EKG results in the report revealed normal sinus rhythm, therefore "RestingECG": "Normal".\\
The maximum heart rate achieved during the treadmill stress test is mentioned as one hundred and fifty beats per minute, therefore "MaxHR": 150.\\
The report states that the patient did not experience chest pain or shortness of breath during the treadmill stress test, so the value is "N" for No, therefore "ExerciseAngina": "N".\\
The old peak value is mentioned in the report as 1.5 mm in leads II, III, and aVF during the treadmill stress test, therefore "Oldpeak": 1.5.\\
The report mentions that the ST-segment slope during the treadmill stress test is downsloping, therefore "ST\_Slope": "Down".\\

Output JSON:
\{'Age': 63, 'Sex': 'M', 'ChestPainType': 'ATA', 'RestingBP': 145, 'Cholesterol': 220, 'FastingBS': 1, 'RestingECG': 'Normal', 'MaxHR': 150, 'ExerciseAngina': 'N', 'Oldpeak': 1.5, 'ST\_Slope': 'Down'\}\\
--------------------------------------------------------------------------------\\
\colorbox{cyan}{\textbf{Medical report: \{Provide medical report\}}}}
\end{tcolorbox}

\subsubsection{Patient treatment classification}
\begin{tcolorbox}[width=\textwidth, title = {Patient treatment classification},
    enhanced, 
    breakable,
    size=small,
    skin first=enhanced,
    skin middle=enhanced,
    skin last=enhanced,
    ]{}
    {\myfont
\colorbox{yellow}{\textbf{Instructions:}}The output JSON should be formatted as a JSON instance that conforms to the JSON schema from Pydantic.\\

As an example, for the schema \{"properties": \{"foo": \{"title": "Foo", "description": "a list of strings", "type": "array", "items": \{"type": "string"\}\}\}, "required": ["foo"]\}\}
the object \{"foo": ["bar", "baz"]\} is a well-formatted instance of the schema. The object \{"properties": \{"foo": ["bar", "baz"]\}\} is not well-formatted.\\

\colorbox{yellow}{\textbf{Here is the output JSON schema:}}\\
```
\{"properties": \{"haematocrit": \{"title": "Haematocrit", "description": "Patient laboratory test result of haematocrit", "type": "number"\}, "haemoglobins": \{"title": "Haemoglobins", "description": "Patient laboratory test result of haemoglobins", "type": "number"\}, "erythrocyte": \{"title": "Erythrocyte", "description": "Patient laboratory test result of erythrocyte", "type": "number"\}, "leucocyte": \{"title": "Leucocyte", "description": "Patient laboratory test result of leucocyte", "type": "number"\}, "thrombocyte": \{"title": "Thrombocyte", "description": "Patient laboratory test result of thrombocyte", "type": "number"\}, "mch": \{"title": "Mch", "description": "Patient laboratory test result of MCH", "type": "number"\}, "mchc": \{"title": "Mchc", "description": "Patient laboratory test result of MCHC", "type": "number"\}, "mcv": \{"title": "Mcv", "description": "Patient laboratory test result of MCV", "type": "number"\}, "age": \{"title": "Age", "description": "Patient age", "type": "integer"\}, "sex": \{"title": "Sex", "description": "Sex of the patient [M,F] where M: Male, F: Female", "type": "string"\}, "source": \{"title": "Source", "description": "In-care patient or out-care patient [in, out] where in: in-care, out: out-care", "type": "string"\}\}\}
```\\

\colorbox{yellow}{\textbf{When generating JSON instance follow this format:}}\\

Medical notes: the input medical notes from which you should extract JSON instance.\\
Reasoning:  give me an explanation of how you assign value for a given key. \\
Thinking step by step for each key before assigning a value to it.\\

Output JSON: The final output JSON should be formatted as a JSON instance that conforms to the output JSON schema above.\\

\colorbox{yellow}{\textbf{Here is an example of a process:}}

Medical notes:\\
A young female individual, currently 12 years old, has come forward with complaints of listlessness and a lack of color. There are no major incidents in her past health record, and she doesn't seem to react adversely to any known substances. A look into her eyes and at her nails shows a lack of color, a finding that is clinically significant. However, no other physical anomalies have been found. Lab studies and the way the individual presents suggest that she may be dealing with anemia.\\
Test results from the lab are as follows:\\
The volume fraction of red blood cells, or the haematocrit, is at a level of 35.1\%, with the typical range for the fairer sex being 36-48\%. This suggests there may be some issues with the blood or it might be anemia. The protein in red blood cells that carries oxygen, haemoglobin, is at a level of 11.8 g/dL, with the norm being 12-16 g/dL, which adds weight to the suspicion of anemia.
The count of erythrocytes, or red blood cells, is 4.65 million cells/mcL, with a normal range of 4-5.2 million cells/mcL. The count of leucocytes, or white blood cells, is 6.3 x 10\^3 cells/mcL, with a normal range of 4-11 x 10\^3 cells/mcL. The thrombocyte, or platelet count, is 310 x 10\^3 cells/mcL, with the normal range being 150-450 x 10\^3 cells/mcL.
}
\end{tcolorbox}

\begin{tcolorbox}[width=\textwidth, notitle,
    size=small,
    ]{}
    {\myfont The mean corpuscular hemoglobin, or MCH, is 25.4 pg. The mean corpuscular hemoglobin concentration, or MCHC, is 33.6 g/dL. The mean corpuscular volume, or MCV, is 75.5 fL, slightly lower than the usual 80-100 fL, hinting at possible microcytic anemia.
Putting the individual's years lived, gender, clinical discoveries, and lab data together, it seems likely that the cause of these symptoms is anemia caused by a lack of iron. For a more definitive answer, more tests, including serum iron, total iron-binding capacity, and ferritin levels, would need to be conducted.\\

\colorbox{orange}{\textbf{Reasoning:}}\\
Haematocrit: The  report states that "The volume fraction of red blood cells, or the haematocrit, is at a level of 35.1\%". Hence, the haematocrit value is 35.1.\\
Haemoglobin: The report states that "The protein in red blood cells that carries oxygen, haemoglobin, is at a level of 11.8 g/dL". Hence, the haemoglobin value is 11.8.\\
Erythrocyte: The report states that "The count of erythrocytes, or red blood cells, is 4.65 million cells/mcL". Hence, the erythrocyte count is 4.65.\\
Leucocyte: The report states that "The count of leucocytes, or white blood cells, is 6.3 x 10\^3 cells/mcL". Hence, the leucocyte count is 6.3.\\
Thrombocyte: The report states that "The thrombocyte, or platelet count, is 310 x 10\^3 cells/mcL". Hence, the thrombocyte count is 310.\\
MCH: The report states that "The mean corpuscular hemoglobin, or MCH, is 25.4 pg". Hence, the MCH value is 25.4.\\
MCHC: The report states that "The mean corpuscular hemoglobin concentration, or MCHC, is 33.6 g/dL". Hence, the MCHC value is 33.6.\\
MCV: The report states that "The mean corpuscular volume, or MCV, is 75.5 fL". Hence, the MCV value is 75.5.\\
Age: The  report refers to "a young female individual, currently stating that she is 12. Hence, the patient's age is 12.\\
Sex: The re-written report refers to "a young female individual". Hence, the sex of the patient is "F".\\

Output JSON:\\
\{"haematocrit": 35.1, "haemoglobin": 11.8, "erythrocyte": 4.65, "leucocyte": 6.3, "thrombocyte": 310, "mch": 25.4, "mchc": 33.6, "mcv": 75.5, "age": 12, "sex": "F", "source": "in" \}\\
--------------------------------------------------------------------------------\\
\colorbox{cyan}{\textbf{Medical report: \{Provide medical report\}}}}
\end{tcolorbox}

\subsubsection{Hepatitis prediction}
\begin{tcolorbox}[width=\textwidth, title = {Hepatitis prediction},
    enhanced, 
    size=small,
    skin first=enhanced,
    skin middle=enhanced,
    skin last=enhanced,
    ]{}
    {\myfont
    \colorbox{yellow}{\textbf{Instructions:}}The output should be formatted as a JSON instance that conforms to the JSON schema below.\\

As an example, for the schema \{"properties": \{"foo": \{"title": "Foo", "description": "a list of strings", "type": "array", "items": \{"type": "string"\}\}\}, "required": ["foo"]\}\}
the object \{"foo": ["bar", "baz"]\} is a well-formatted instance of the schema. The object \{"properties": \{"foo": ["bar", "baz"]\}\} is not well-formatted.\\

\colorbox{yellow}{\textbf{Here is the output schema:}}\\
```
\{"properties": \{"age": \{"title": "Age", "description": "Age of the patient in years", "type": "integer"\}, "sex": \{"title": "Sex", "description": "Sex of the patient [f, m] ('f'=female, 'm'=male)", "type": "string"\}, "ALB": \{"title": "Alb", "description": "Amount of albumin in patient's blood", "type": "number"\}, "ALP": \{"title": "Alp", "description": "Amount of alkaline phosphatase in patient's blood", "type": "number"\}, "ALT": \{"title": "Alt", "description": "Amount of alanine transaminase in patient's blood", "type": "number"\}, "AST": \{"title": "Ast", "description": "Amount of aspartate aminotransferase in patient's blood", "type": "number"\}, "BIL": \{"title": "Bil", "description": "Amount of bilirubin in patient's blood", "type": "number"\}, "CHE": \{"title": "Che", "description": "Amount of cholinesterase in patient's blood", "type": "number"\}, "CHOL": \{"title": "Chol", "description": "Amount of cholesterol in patient's blood", "type": "number"\}, "CREA": \{"title": "Crea", "description": "Amount of creatine in patient's blood", "type": "number"\}, "GGT": \{"title": "Ggt", "description": "Amount of gamma-glutamyl transferase in patient's blood", "type": "number"\}, "PROT": \{"title": "Prot", "description": "Amount of protein in patient's blood", "type": "number"\},\}\}\}
```\\

\colorbox{yellow}{\textbf{When generating JSON instance follow this format:}}\\

Medical notes: the input medical notes from which you should extract JSON instance
Reasoning:  give me an explanation of how you assign value for a given key. Thinking step by step for each key before assigning a value to it.
Output JSON: The final output JSON should be formatted as a JSON instance that conforms to the output JSON schema above.\\

\colorbox{yellow}{\textbf{Here is an example of a process:}}\\
Medical notes:\\
An individual male, having completed 32 solar cycles, underwent a series of tests aimed at examining hepatic function. An absence of noteworthy historical and clinical data was noted for this individual. As per the scrutinized lab results, his hepatic functions seem to be running smoothly.\\
The concentration of a certain protein - albumin, to be exact - was noted to be 38.5 g/L, comfortably residing within the accepted range of 35-55 g/L. This could be a sign of sufficient protein production by the liver. The alkaline phosphatase, or ALP, measured at 70.3 U/L, is also within the typical parameters (40-130 U/L), hinting at the lack of severe hepatic or skeletal disorders.
Additional indicators, namely alanine transaminase (ALT) and aspartate aminotransferase (AST), clocked in at 18.0 U/L and 24.7 U/L respectively, both within the usual limits (7-56 U/L for ALT, and 10-40 U/L for AST). This further bolsters the suggestion of no significant hepatic or muscular impairment. Bilirubin, or BIL, another liver health marker, was found to be 3.9 mmol/L, a level that falls within the normal range (1.2-17.1 mmol/L), suggesting no significant hepatic disorders or anemia. Another chemical, cholinesterase (CHE), was found to be 11.17 U/L, again within the normal limits (5.3-12.9 U/L), which implies no significant liver issues.
}
\end{tcolorbox}

\begin{tcolorbox}[width=\textwidth, notitle,
    size=small,
    ]{}
    {\myfont
    The cholesterol or CHOL level, a marker of lipid health, was noted to be 4.8 mmol/L, fitting within the normal range (3.6-7.8 mmol/L), suggesting a good lipid profile. Creatinine or CREA, a marker of kidney health, was found to be 74.0 mmol/L, a value within the normal range (62-106 mmol/L), indicating no renal issues.\\
Finally, gamma-glutamyl transferase (GGT) and total protein (PROT), with values of 15.6 U/L and 76.5 g/L respectively, were both within their normal ranges (9-64 U/L for GGT, and 60-83 g/L for PROT). This implies no significant damage to the liver or bile duct and adequate protein production by the liver.
In summary, upon thorough investigation of the lab data, it can be inferred that the liver is operating satisfactorily with no signs of damage or dysfunction. Lipid and kidney health also appear to be maintained within the normal parameters. There is no necessity for additional interventions at this stage.\\

\colorbox{orange}{\textbf{Reasoning:}}\\
'age' key, the text mentions that the individual has 'completed 32 solar cycles', which is a creative way of saying that the patient is 32 years old. Therefore, "age": 32.\\
'sex', the report mentions 'an individual of the male gender', which means the patient is male. Therefore, "sex": "m".\\
'ALB' key stands for the albumin level, which the report states as '38.5 g/L'. Therefore, "ALB": 38.5.\\
'ALP', representing alkaline phosphatase, the report mentions it as '70.3 U/L'. Therefore, "ALP": 70.3.\\
'ALT' and 'AST', standing for alanine transaminase and aspartate aminotransferase respectively, the report mentions them as '18.0 U/L' and '24.7 U/L' respectively. Therefore, "ALT": 18.0 and "AST": 24.7.\\
'BIL' key represents the bilirubin level, which the report states as '3.9 mmol/L'. Therefore, "BIL": 3.9.\\
'CHE', indicating cholinesterase, the report provides a figure of '11.17 U/L'. Therefore, "CHE": 11.17.\\
'CHOL', standing for cholesterol, the report mentions it as '4.8 mmol/L'. Therefore, "CHOL": 4.8.\\
'CREA' key, representing creatinine, is mentioned in the report as '74.0 mmol/L'. Therefore, "CREA": 74.0.\\
'GGT' and 'PROT', standing for gamma-glutamyl transferase and total protein respectively, are stated as '15.6 U/L' and '76.5 g/L' respectively. Therefore, "GGT": 15.6 and "PROT": 76.5.\\

Output JSON:\\
\{'Age': 32, 'Sex': 'm', 'ALB': 38.5, 'ALP': 70.3, 'ALT': 18.0, 'AST': 24.7, 'BIL': 3.9, 'CHE': 11.17, 'CHOL': 4.8, 'CREA': 74.0, 'GGT': 15.6, 'PROT': 76.5\}\\
--------------------------------------------------------------------------------\\
\colorbox{cyan}{\textbf{Medical report: \{Provide medical report\}}}}
\end{tcolorbox}

\subsubsection{Stroke prediction}

\begin{tcolorbox}[width=\textwidth, title = {Stroke prediction},
    enhanced, 
    size=small,
    skin first=enhanced,
    skin middle=enhanced,
    skin last=enhanced,
    ]{}
    {\myfont
    \colorbox{yellow}{\textbf{Instructions:}}The output should be formatted as a JSON instance that conforms to the JSON schema below.\\

As an example, for the schema \{"properties": \{"foo": \{"title": "Foo", "description": "a list of strings", "type": "array", "items": \{"type": "string"\}\}\}, "required": ["foo"]\}\}
the object \{"foo": ["bar", "baz"]\} is a well-formatted instance of the schema. The object \{"properties": \{"foo": ["bar", "baz"]\}\} is not well-formatted.\\

\colorbox{yellow}{\textbf{Here is the output schema:}}\\
```
\{"properties": \{"choose\_your\_gender": \{"title": "Choose Your Gender", "description": "Gender of the student [Female, Male]", "type": "string"\}, "age": \{"title": "Age", "description": "Age of the student", "type": "number"\}, "what\_is\_your\_course": \{"title": "What Is Your Course", "description": "Course the student is enrolled in", "type": "string"\}, "your\_current\_year\_of\_study": \{"title": "Your Current Year Of Study", "description": "Current year of study [year 1, year 2, year 3, year 4]", "type": "string"\}, "what\_is\_your\_cgpa": \{"title": "What Is Your Cgpa", "description": "Current CGPA range of the student [0 - 1.99, 2.00 - 2.49, 2.50 - 2.99, 3.00 - 3.49, 3.50 - 4.00]", "type": "string"\}, "marital\_status": \{"title": "Marital Status", "description": "Marital status of the student [yes, no]", "type": "string"\}, "do\_you\_have\_depression": \{"title": "Do You Have Depression", "description": "Indicates if the student has depression [yes, no]", "type": "string"\}, "do\_you\_have\_anxiety": \{"title": "Do You Have Anxiety", "description": "Indicates if the student has anxiety [yes, no]", "type": "string"\}, "do\_you\_have\_panic\_attack": \{"title": "Do You Have Panic Attack", "description": "Indicates if the student has panic attacks [yes, no]", "type": "string"\}, "did\_you\_seek\_any\_specialist\_for\_treatment": \{"title": "Did You Seek Any Specialist For Treatment", "description": "Indicates if the student has sought treatment from a specialist [yes, no]", "type": "string"\}\}\}\\
```

\colorbox{yellow}{\textbf{When generating JSON instance follow this format:}}\\

Medical notes: the input medical notes from which you should extract JSON instance.\\
Reasoning:  \\
Give me an explanation of how you assign value for a given key. Thinking step by step for each key before assigning a value to it.\\
Output JSON: The final output JSON should be formatted as a JSON instance that conforms to the output JSON schema above.\\

\colorbox{yellow}{\textbf{Here is an example of a process:}}\\

Medical report:\\
Subject: 49-year-old Female Patient.\\

I. Patient History
A 49-year-old married female patient, employed in the private sector and residing in an urban area, presents for evaluation. She reports no history of hypertension or heart disease.\\

II. Clinical Findings\\
A. No hypertension\\
B. No heart disease\\
}
\end{tcolorbox}

\begin{tcolorbox}[width=\textwidth, notitle,
    enhanced, 
    size=small,
    ]{}
    {\myfont
    III. Diagnostic Laboratory Test Results\\
A. Average Glucose Level: 171.23 mg/dL (elevated, normal range: 70-140 mg/dL)\\
B. Body Mass Index (BMI): 34.4 kg/m2 (classified as obesity, normal range: 18.5-24.9 kg/m2)\\

IV. Risk Factors\\
A. Smoking status: Active smoker\\

V. Comments on General Presentation\\
The patient exhibits an elevated average glucose level and a BMI within the obesity range. She has no prior history of hypertension or heart disease. However, the patient actively smokes, a known risk factor for various health complications.\\

VI. Management\\
A. Lifestyle Recommendations\\
Healthy diet: Encourage the patient to consume a balanced diet, rich in fruits, vegetables, whole grains, lean protein, and healthy fats.
Physical activity: Encourage the patient to engage in moderate-intensity aerobic exercise for at least 150 minutes per week or vigorous-intensity aerobic exercise for at least 75 minutes per week, in addition to muscle-strengthening activities on two or more days per week.
Smoking cessation: Recommend smoking cessation programs and resources to help the patient quit smoking.\\

B. Monitoring\\
Regular check-ups: Schedule periodic appointments to monitor the patient's glucose levels, blood pressure, and cholesterol.\\

VII. Opinion\\
Considering the patient's elevated average glucose level, obesity, and smoking habit, her risk for future health complications, including diabetes, hypertension, and heart disease, is significantly increased. Close monitoring and adherence to recommended lifestyle changes are crucial to minimize the risk of further health issues and improve overall health outcomes.\\

Output JSON:\\
\{"gender": "Female", "age": 49.0, "hypertension": 0, "heart\_disease": 0, "ever\_married": "Yes", "work\_type": "Private", "Residence\_type": "Urban", "avg\_glucose\_level": 171.23, "bmi": 34.4, "smoking\_status": "smokes"\}\\
--------------------------------------------------------------------------------\\
\colorbox{cyan}{\textbf{Medical report: \{Provide medical report\}}}}
\end{tcolorbox}

\pagebreak

\subsubsection{Psychologist notes}
\begin{tcolorbox}[width=\textwidth, title = {Psychologist notes},
    enhanced, 
    breakable,
    size=small,
    skin first=enhanced,
    skin middle=enhanced,
    skin last=enhanced,
    ]{}
    {\myfont
    \colorbox{yellow}{\textbf{Instructions:}}The output should be formatted as a JSON instance that conforms to the JSON schema below.\\

As an example, for the schema \{"properties": \{"foo": \{"title": "Foo", "description": "a list of strings", "type": "array", "items": \{"type": "string"\}\}\}, "required": ["foo"]\}\}
the object \{"foo": ["bar", "baz"]\} is a well-formatted instance of the schema. The object \{"properties": \{"foo": ["bar", "baz"]\}\} is not well-formatted.\\

\colorbox{yellow}{\textbf{Here is the output schema:}}\\
```
\{"properties": \{"choose\_your\_gender": \{"title": "Choose Your Gender", "description": "Gender of the student [Female, Male]", "type": "string"\}, "age": \{"title": "Age", "description": "Age of the student", "type": "number"\}, "what\_is\_your\_course": \{"title": "What Is Your Course", "description": "Course the student is enrolled in", "type": "string"\}, "your\_current\_year\_of\_study": \{"title": "Your Current Year Of Study", "description": "Current year of study [year 1, year 2, year 3, year 4]", "type": "string"\}, "what\_is\_your\_cgpa": \{"title": "What Is Your Cgpa", "description": "Current CGPA range of the student [0 - 1.99, 2.00 - 2.49, 2.50 - 2.99, 3.00 - 3.49, 3.50 - 4.00]", "type": "string"\}, "marital\_status": \{"title": "Marital Status", "description": "Marital status of the student [yes, no]", "type": "string"\}, "do\_you\_have\_depression": \{"title": "Do You Have Depression", "description": "Indicates if the student has depression [yes, no]", "type": "string"\}, "do\_you\_have\_anxiety": \{"title": "Do You Have Anxiety", "description": "Indicates if the student has anxiety [yes, no]", "type": "string"\}, "do\_you\_have\_panic\_attack": \{"title": "Do You Have Panic Attack", "description": "Indicates if the student has panic attacks [yes, no]", "type": "string"\}, "did\_you\_seek\_any\_specialist\_for\_treatment": \{"title": "Did You Seek Any Specialist For Treatment", "description": "Indicates if the student has sought treatment from a specialist [yes, no]", "type": "string"\}\}\}
```\\

\colorbox{yellow}{\textbf{When generating JSON instance follow this format:}}\\
Medical notes: the input medical notes from which you should extract JSON instance.\\
Reasoning:  give me an explanation of how you assign value for a given key. Thinking step by step for each key before assigning a value to it.\\
Output JSON: The final output JSON should be formatted as a JSON instance that conforms to the output JSON schema above.\\

\colorbox{yellow}{\textbf{Here is an example of a process:}}\\
Medical notes:\\
In the middle of a summer day on the seventh of August, 2020, precisely at four past midday, an encounter took place with a scholar who is in his early twenties. This young man is pursuing knowledge in the field of Islamic Studies and has already completed a full academic year, making him a second-year apprentice of knowledge. His academic performance has been commendable, as he's managed to secure a Cumulative Grade Point Average (CGPA) that floats between the range of 3.00 and 3.49.\\

He made it known that he currently has no spouse, a state which inherently may mold the structure of his social interactions as well as the challenges he grapples with. This scholar divulged worries related to his mental well-being, with a specific reference to anxiety. Interestingly, he denied experiencing depressive episodes or panic-filled incidents, this might provide us with a more streamlined avenue to address his anxiety.\\

}
\end{tcolorbox}

\begin{tcolorbox}[width=\textwidth, notitle,
    enhanced, 
    size=small,
    skin first=enhanced,
    skin middle=enhanced,
    skin last=enhanced,
    ]{}
    {\myfont
    Past attempts at seeking professional guidance or treatment for his concerns were negated. The information gleaned from this interaction suggests the possibility that the scholar may stand to gain from more thorough evaluations and tailored interventions to navigate his anxiety more effectively.\\

To this end, a referral to a proficient hand in the field of mental health, like a clinical psychologist or counselor, might prove useful. In tandem with this, the scholar might find solace in engaging with methods designed to manage stress and relaxation exercises, with the aim of taking control of his anxiety symptoms.\\

\colorbox{orange}{\textbf{Reasoning:}}\\
Choose Your Gender: The medical notes refer to the individual as a "young man" and "he," indicating that the person identifies as male.\\
Age: The notes refer to the individual as being "in his early twenties." Given the lack of a specific age, we will assign the value of None. Therefore, "age": None.\\
What\_is\_your\_course: It's mentioned that the individual is studying "Islamic Studies." Therefore, "what\_is\_your\_course": "Islamic Studies".\\
Your\_current\_year\_of\_study: The notes state that the individual "has already completed a full academic year," indicating that he's currently in his second year of study. Therefore, "your\_current\_year\_of\_study": "year 2".\\
What\_is\_your\_cgpa: The notes mention a CGPA "between the range of 3.00 and 3.49." Therefore, "what\_is\_your\_cgpa": "3.00 - 3.49".\\
Marital\_status: The individual stated he "has no spouse," indicating he is not married. Therefore, "marital\_status": "no".\\
Do\_you\_have\_depression: The notes specifically state that the individual denied "experiencing depressive episodes," indicating he does not have depression. Therefore, "do\_you\_have\_depression": "no".\\
Do\_you\_have\_anxiety: The individual has expressed concerns about his mental well-being, specifically mentioning anxiety. Therefore, "do\_you\_have\_anxiety": "yes".\\
Do\_you\_have\_panic\_attack: The notes state that the individual denied "panic-filled incidents," indicating he does not have panic attacks. Therefore, "do\_you\_have\_panic\_attack": "no".\\
Did\_you\_seek\_any\_specialist\_for\_treatment: It's mentioned in the notes that the individual has not sought professional help for his concerns in the past. Therefore, "did\_you\_seek\_any\_specialist\_for\_treatment": "no".\\

Output JSON:\\
\{"choose\_your\_gender": "Male", 
"age": None, 
"what\_is\_your\_course": "Islamic Studies", 
"your\_current\_year\_of\_study": "year 2", 
"what\_is\_your\_cgpa": "3.00 - 3.49", 
"marital\_status": "no", 
"do\_you\_have\_depression": "no", 
"do\_you\_have\_anxiety": "yes", 
"do\_you\_have\_panic\_attack": "no 
\}\\
--------------------------------------------------------------------------------\\
\colorbox{cyan}{\textbf{Medical report: \{Provide medical report\}}}}
\end{tcolorbox}

\pagebreak

\subsection{TEMED-LLM {\myfont dtree} Visualization}
\label{app:dtree-vis}

\begin{figure}
	\center
	\includegraphics[angle=0, width=1\textwidth]{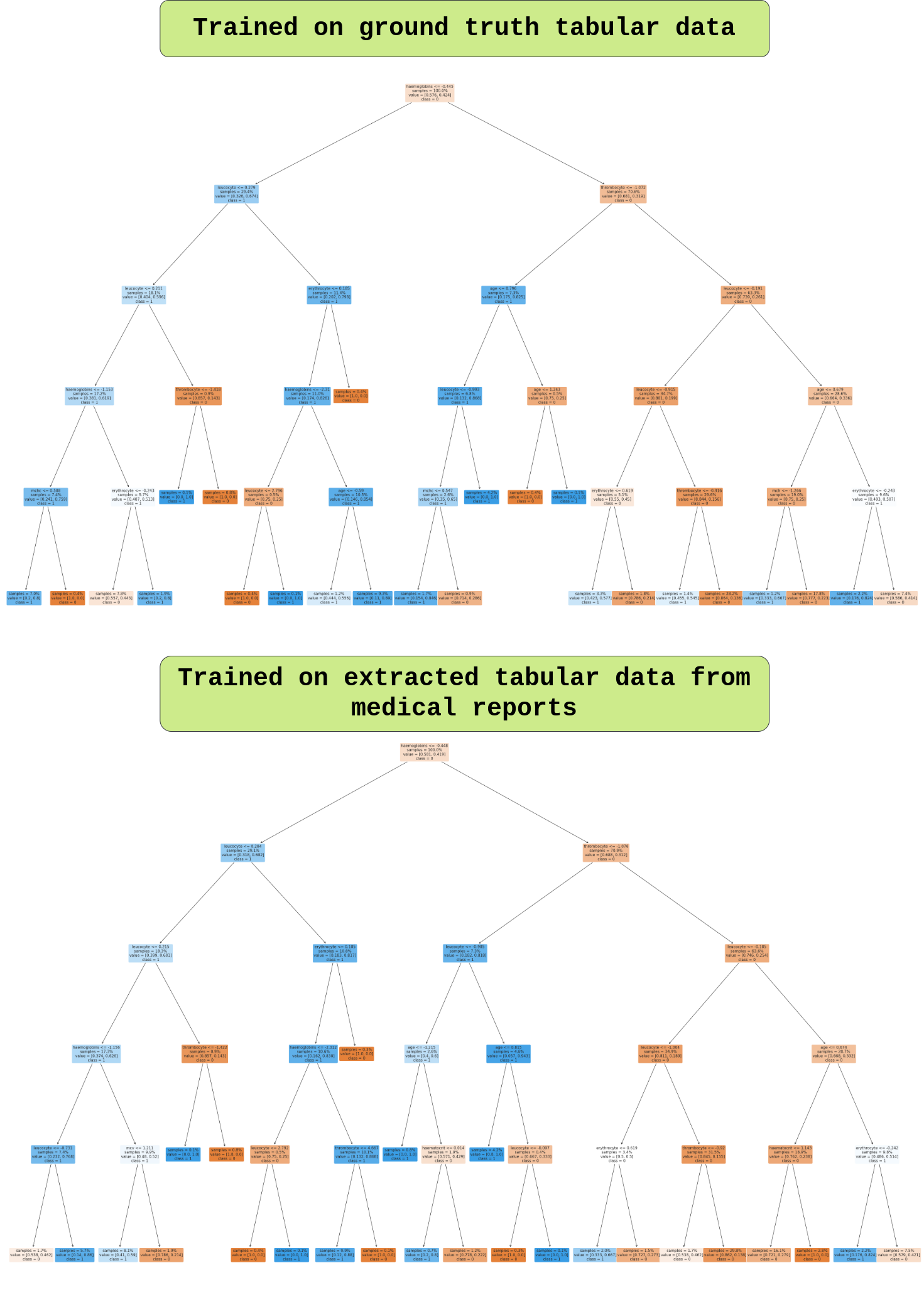}
	\captionsetup{justification=centering}
	\caption{Comparison of two decision trees trained on the Patient treatment classification dataset: the first one utilizing ground truth tabular data and the second one using tabular data extracted from medical reports by the TEMED-LLM methodology.The color intensity on the nodes of the decision tree signifies the probability that instances of either class 0 (depicted in orange) or class 1 (shown in blue) dominate that node.}
	\label{fig:dtree}
\end{figure}

Figure~\ref{fig:dtree} visualizes two decision trees, both of which were trained on the Patient treatment classification dataset. The first decision tree is trained on ground truth tabular data, while the second one uses tabular data extracted from medical reports, employing the TEMED-LLM methodology.
Upon inspection, it's clear that the decision trees exhibit striking similarities. They share a similar structure, and only a minimal amount of differences can be observed between them. This comparison further illustrates
the conclusions drawn in section~\ref{Sec:interpretability}. 

\subsection{Details of used ML models and their setups} 
\label{app:details}

In this section, we present details and model setups of text classifiers and tabular data models.

\subsubsection{Text classifiers}
In our study, we employed four different natural language processing (NLP) text classification models: {\myfont 'xlm-roberta-large'}, {\myfont 'roberta-base'}, {\myfont 'bert-base-cased'}, and {\myfont 'dmis-lab/biobert-v1.1'}. Each model's configuration and hyperparameters are described below.

The {\myfont 'xlm-roberta-large'} leverages the XLM-RoBERTa architecture, known for its strong performance on multilingual and cross-lingual tasks. It consists of 24 hidden layers and 16 attention heads, with a hidden size of 1024 and a maximum position embedding of 514. It returns dictionaries by default and uses GELU as the hidden activation function. The model employs an absolute position embedding type and has a vocabulary size of 250,002. For training, we used a learning rate of 2e-05, a per-device train batch size of 16, and set the number of training epochs to 3. Weights decayed at a rate of 0.01.

Based on the RoBERTa model, {\myfont 'roberta-base'} variant has 12 hidden layers and 12 attention heads. It utilizes a hidden size of 768 and a maximum position embedding of 514. Like the XLM-RoBERTa-Large model, it returns dictionaries by default, uses GELU as the hidden activation function, and employs an absolute position embedding type. The vocabulary size of the RoBERTa-Base model is 50,265. Its hyperparameters for training are identical to those of the XLM-RoBERTa-Large model.

The {\myfont 'bert-base-cased'} operates on the BERT architecture, which pioneered transformer-based language models. It has a configuration similar to RoBERTa-Base, with 12 hidden layers and 12 attention heads. The hidden size and maximum position embeddings are also identical, at 768 and 512 respectively. However, the vocabulary size is slightly smaller, at 28,996. The training hyperparameters remain the same as the previous two models.

BioBERT ({\myfont 'dmis-lab/biobert-v1.1'}) is a domain-specific variant of BERT that is pre-trained on large-scale biomedical corpora. Similar to BERT-Base-Cased, this model has 12 hidden layers and 12 attention heads, a hidden size of 768, and a maximum position embedding of 512. The vocabulary size matches that of BERT-Base-Cased, at 28,996. BioBERT's training hyperparameters are identical to those of the previous models.

The {\myfont 'set-fit-mpnetv2'} model is designed for text processing tasks. It comprises 12 hidden layers with 12 attention heads and a hidden size of 768 dimensions. The vocabulary size is 30,527 tokens and it uses the GeLU activation function. Training involves a learning rate of 2e-05, training and evaluation batch sizes of 16, and 3 epochs.

During training the models were set to perform evaluations and save the model at the end of each epoch. The best model was loaded at the end of the training process.

\subsubsection{ML tabular models}
In this study, we used three ML models for tabular data: logistic regression ({\myfont logreg}), decision tree ({\myfont dtree}), and XGBoost ({\myfont xgboost}). We provide specific setups below.

For the ({\myfont logreg}), we use {\myfont scikit-learn}\footnote{\url{https://scikit-learn.org/}} implementation. The 'C' parameter, which controls the inverse of the regularization strength, was chosen from a range of values [0.001, 0.01, 0.1, 1, 10, 100]. The model used an L2 penalty (ridge) with the solver 'lbfgs' for optimization. We ensured that the model includes an intercept in the decision function. The model was allowed to converge to a solution within a maximum of 100 iterations.

The ({\myfont dtree}), we use {\myfont scikit-learn} implementation. The model was set up with the Gini impurity criterion, indicating the use of the Gini impurity. The maximum depth of the tree was explored with values [3, 4, 5], while the minimum number of samples required to split an internal node was picked from [2, 3, 4, 5, 7, 10]. At each node, the best split was chosen (i.e., the split that results in the maximum reduction of the impurity criterion).

Finally, for the {\myfont xgboost} model we use {\myfont XGBoost}\footnote{\url{https://xgboost.readthedocs.io/}} package. The number of gradient-boosted trees was chosen from [50, 100, 200] and the learning rate was selected from [0.01, 0.1, 0.3]. The rest of the hyperparameters was set to default values.

All the models were trained without considering any class weights, implying that all classes have equal importance in model learning. Hyperparameters were tuned using a grid search method. 

\end{appendix}
%%%%%%%%%%%%%%%%%%%%%%%%%%%%%%%%%%%%%%%%%%%%%%%%%%%%%%%%%%%%

\end{document}